\documentclass{article}

\usepackage{PRIMEarxiv}
\usepackage[utf8]{inputenc} % allow utf-8 input
\usepackage[T1]{fontenc}    % use 8-bit T1 fonts
\usepackage{hyperref}       % hyperlinks
\usepackage{url}            % simple URL typesetting
\usepackage{booktabs}       % professional-quality tables
\usepackage{amsfonts}       % blackboard math symbols
\usepackage{nicefrac}       % compact symbols for 1/2, etc.
\usepackage{microtype}      % microtypography
\usepackage{lipsum}
\usepackage{fancyhdr}       % header
\usepackage{graphicx}       % graphics
\graphicspath{{media/}}     % organize your images and other figures under media/ folder
\usepackage{xspace}
\usepackage{amsmath,amssymb,amsfonts}
\usepackage{nomencl}
\usepackage[utf8]{inputenc}

\usepackage{tabularx} % Allows for table column width to be specified
\usepackage{longtable} % For tables that span multiple pages
\usepackage{geometry} % Adjust page margins
\title{Deep Reinforcement Learning in Autonomous Car Path Planning and Control: A Survey
}

\author{
  Yiyang Chen$^1$$^2$, Chao Ji$^1$$^3$, Yunrui Cai$^1$$^2$, Tong Yan$^1$, Bo Su$^{1*}$\\
    $^1$China North Artificial Intelligence and Innovation Research Institute, Beijing, 100072, China \\
    $^2$Department of Precision Instruments, Tsinghua University, Beijing, China \\
    $^3$Department of Automation, Tsinghua University, Beijing, China\\
  \texttt{Email:chenyy22}@mails.tsinghua.edu.cn}

\begin{document}
\maketitle

\begin{abstract}
Combining data-driven applications with control systems plays a key role in recent Autonomous Car research. This thesis offers a structured review of the latest literature on Deep Reinforcement Learning (DRL)  within the realm of autonomous vehicle Path Planning and Control. It collects a series of DRL methodologies and algorithms and their applications in the field, focusing notably on their roles in trajectory planning and dynamic control. In this review, we delve into the application outcomes of DRL technologies in this domain. By summarizing these literatures, we highlight potential challenges, aiming to offer insights that might aid researchers engaged in related fields. 

\end{abstract}

\keywords{ Deep Reinforcement Learning (DRL) \and Autonomous Vehicles \and Path Planning \and Automotive Control \and path tracking}

\section{Introduction}  \label{sec: intro}

As autonomous driving technology rapidly advances, its potential to relieve drivers, enhance traffic efficiency, reduce energy consumption, and improve road safety is increasingly being recognized\cite{yurtsever2020survey}. At present, advancements in autonomous vehicle control technologies are chiefly derived from the integration of Advanced Driver Assistance Systems (ADAS), including Adaptive Cruise Control (ACC), Lane Keeping Assistance Systems, and Lane Departure Warning technologies, which have been implemented in a variety of commercial electric vehicles. Projects such as Google's Waymo and Baidu's Apollo have advanced towards commercial operations, achieving autonomous driving capabilities and launching unmanned vehicle rental services in designated areas.

The control framework of autonomous vehicles fundamentally encompasses three tiers: perception, planning, and control, with Figure 1 \cite{stano2023model} depicting the comprehensive architecture of autonomous driving systems. The perception layer is tasked with the accurate perception and processing of measurement data to produce dependable state estimates essential for precise localization and environmental recognition. The planning layer is dedicated to task planning, behavioral planning, and path planning, ensuring the vehicle is capable of making judicious decisions within intricate environments. The control layer concentrates on sustaining vehicle stability and adhering precisely to the pre-determined trajectory. The success in perception and prediction is largely attributable to substantial advancements in the field of machine vision in recent years. However, the planning and control segments depend on developers setting parameters for the hierarchical controller and conducting fine-tuning through simulations and on-site testing, a traditional methodology that presents distinct limitations, particularly in parameter adjustment and adaptation to new settings, proving to be both time-consuming and inefficient. Moreover, given the highly nonlinear characteristics of the driving process, control strategies dependent on the linearization of vehicle models or other algebraic solutions encounter challenges regarding implementation and scalability\cite{artunedo2024lateral}.

\begin{figure}[h!]
    \centering
    \includegraphics[width=\linewidth]{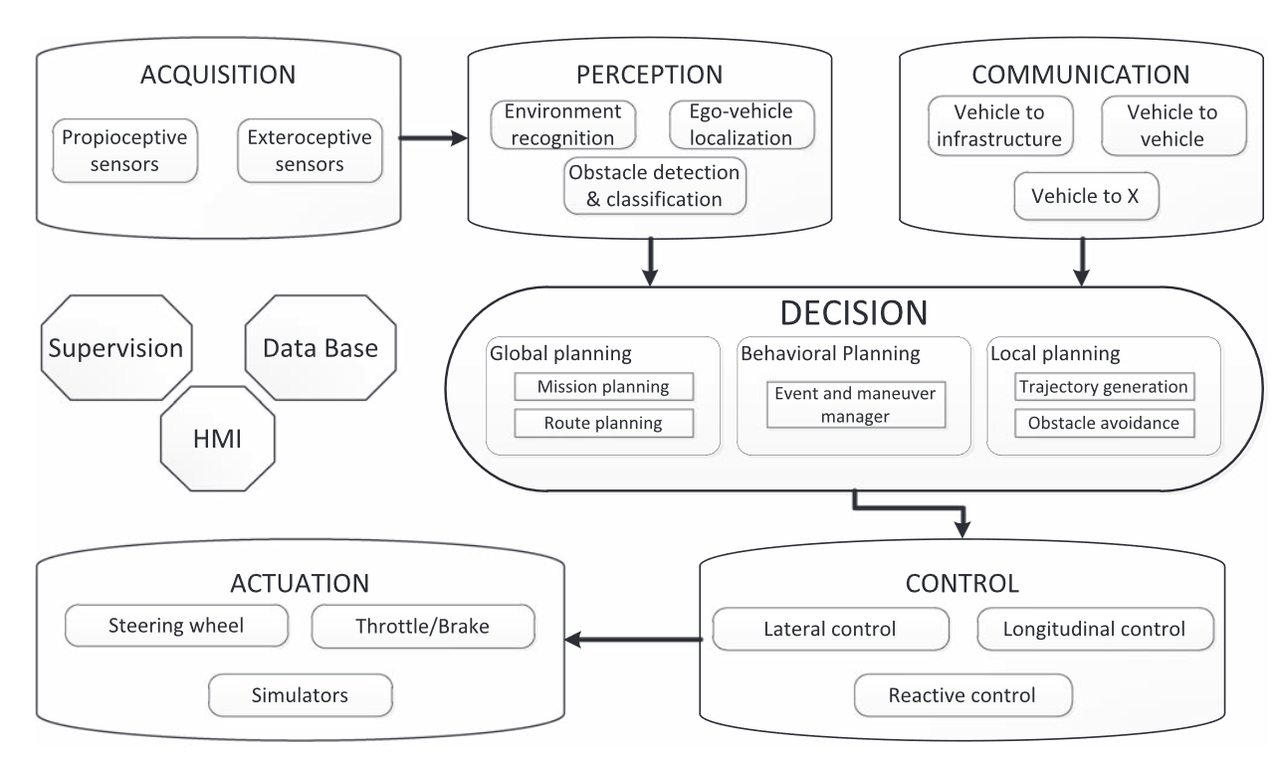}
    \caption{\textbf{Control architecture for automated vehicles\cite{stano2023model}.}}
    \label{fig:control_architecture}
\end{figure}

Reinforcement Learning (RL) is founded upon the Markov Decision Process (MDP), enabling the continual execution of actions in response to a system of rewards and punishments. The integration of reinforcement learning with deep learning, termed Deep Reinforcement Learning (DRL), signifies the cutting edge in learning frameworks for control systems. Deep Reinforcement Learning (DRL) is adept at approximating highly non-linear functions from complex datasets, addressing the complex control challenges within various subfields of autonomous driving, such as behavioral decision-making, energy management, and traffic flow control. Within the realm of behavioral decision-making, advanced planning and behavioral prediction employ Deep Reinforcement Learning (DRL) and Deep Inverse Reinforcement Learning to navigate and make decisions within dynamic and uncertain environments\cite{you2019advanced}. Furthermore, the application of multi-agent DRL in the management and optimization of traffic flows presents a promising avenue for addressing and mitigating traffic congestion issues, thus enhancing overall traffic efficiency. This innovative application of DRL not only underscores its versatility but also demonstrates its potential in orchestrating complex traffic dynamics, paving the way for the creation of more streamlined and efficient transportation systems\cite{haydari2020deep}.

In the realm of motion planning and control, the dynamic and uncertain nature of driving environments necessitates sophisticated strategies capable of adapting to real-time changes while ensuring optimal path selection. The integration of Deep Reinforcement Learning (DRL) has been demonstrated to effectively navigate these complexities, providing evolving solutions through continual environmental learning\cite{aradi2020survey,claussmann2019review,kuutti2020survey}. For instance, in the context of autonomous vehicle planning problems grounded in stochastic MDPs, DRL facilitates the simulation of interactions between vehicles and their environment, learning optimal strategies like overtaking and tailgating. This approach simultaneously takes into account the vehicle's status, destination, and potential obstacles, thus ensuring the accuracy of path planning and passenger comfort. While Deep Reinforcement Learning (DRL) presents significant advantages, the challenges of effectively learning from sparse, high-dimensional sensory inputs and ensuring safety and robustness in actual driving scenarios persist.

This paper centers on local path planning and motion control within autonomous driving, examining the merits and current landscape of Reinforcement Learning (RL) and Deep Reinforcement Learning (DRL) by gathering, assessing, and synthesizing pertinent literature to underscore significant accomplishments and burgeoning research avenues. Concurrently, through meticulous analysis, this work aims to illuminate the pivotal role of RL in the advancement of autonomous driving technologies, particularly in the realm of precise trajectory planning and control, while also identifying unresolved challenges and prospective research directions within the current domain.

To ascertain the pivotal articles for this investigation, a systematic review procedure as illustrated in Figure 2 was utilized. Initially, through keyword-based topic searches on the Web of Science database, LLM tools, and word embedding techniques, a preliminary filtration was performed, retaining about 500 articles that integrate Reinforcement Learning (RL) with autonomous driving trajectory control. Subsequently, a preliminary categorization of the research domains was conducted based on the abstracts of each article, retaining those pertinent to trajectory planning and motion control, thus reducing the count to 180 articles. Within the selected cohort of 180 articles, a comprehensive review of the full texts was undertaken, selecting 45 of the most recent and pertinent articles. In this review article, a recapitulation of the latest developments in the application of RL to trajectory planning and control is provided, based on these 45 articles. Furthermore, an additional 50+ reference articles were incorporated to encompass descriptions of the most advanced RL methodologies and delineate issues related to autonomous driving trajectory planning and control.

\begin{figure}[h!]
    \centering
    \includegraphics[width=\linewidth]{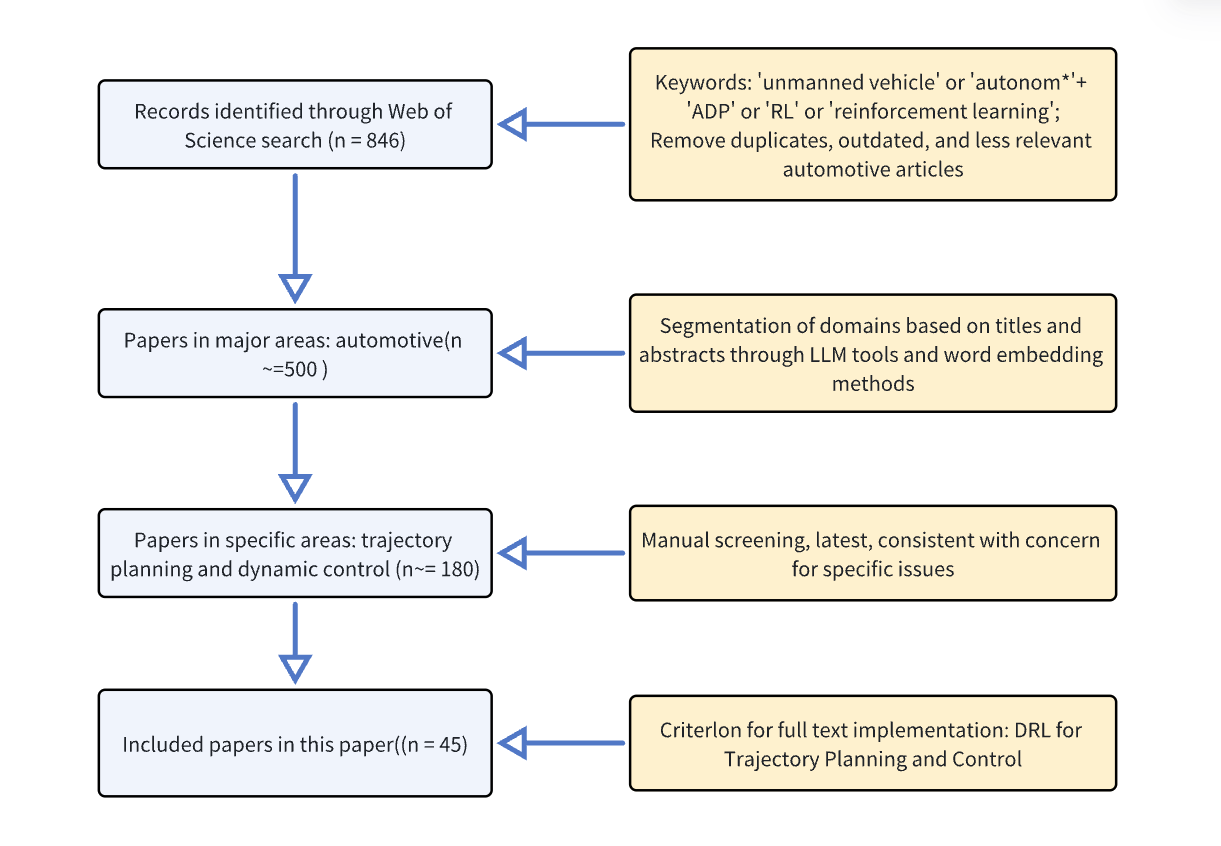}
    \caption{\textbf{Methodology of screening and selecting papers in this review paper.}}
    \label{fig:screening and selecting papers}
\end{figure}

Chapter 2 of this paper will succinctly introduce mainstream reinforcement learning methods and their applicability across a spectrum of autonomous driving challenges. Chapter 3 delves into trajectory planning issues, encompassing the challenges of established methodologies and detailing the application of reinforcement learning methods tailored to specific scenarios. Chapter 4 explores lateral and longitudinal control challenges, along with the implementation of reinforcement learning in addressing these issues. Chapter 5 presents the recently popular end-to-end approaches, wherein reinforcement learning agents integrate path planning with motion control. Finally, a summary and discussion of the current literature on DRL methods will be provided, highlighting the challenges and potential future research directions for DRL's further application in the sub-domains of trajectory planning and motion control.

\section{Overview of DRL in Autonomous Driving}

\subsection{Introduction to Reinforcement Learning Algorithms}

RL is a subset of machine learning in which an agent learns to make decisions through the execution of actions within an environment to attain a specified objective. The core of RL lies in the interaction between the agent and its environment, facilitated through a trial-and-error process. The agent receives feedback via rewards, directing it towards a strategy that optimizes the accumulation of rewards over time. This learning paradigm is distinguished by its focus on learning the optimal action selection strategy, without requiring a pre-defined environmental model, rendering it particularly apt for complex and dynamically uncertain applications like autonomous driving.

Within the context of autonomous driving, particularly regarding motion planning and control, the Markov Decision Process (MDP) offers a foundational framework for modeling the decision-making process. The MDP is delineated by its states, actions, transition probabilities, and rewards. States can encompass a broad array of information, including the vehicle’s current speed, position, direction, and the surrounding environmental conditions, like the proximity to other vehicles. Actions denote the comprehensive set of operations a vehicle can execute, such as accelerating, braking, or turning at various angles. The MDP framework presupposes that future states are contingent solely upon the current state and action, adhering to the Markov property. This assumption simplifies the complexity involved in forecasting future scenarios based on the entirety of historical data.

\begin{figure}[h!]
    \centering
    \includegraphics[width=0.6\linewidth]{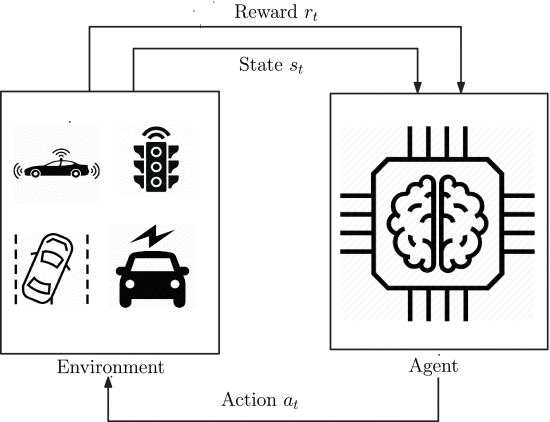}
    \caption{\textbf{MDP for RL loop\cite{haydari2020deep}.}}
    \label{fig:MDP}
\end{figure}

The reward function plays a pivotal role in Reinforcement Learning (RL), as it directly influences the agent's behavior by incentivizing actions that guide towards desired outcomes. Within the context of autonomous driving, the reward function can be crafted to prioritize aspects such as safety, efficiency, and compliance with traffic regulations. For instance, rewards can be allocated for behaviors such as maintaining a safe distance from other vehicles, reaching the destination within a designated time frame, or minimizing fuel consumption. Conversely, penalties might be imposed for actions deemed dangerous, such as speeding or deviating from the planned route. The state transition function, or the environment within the MDP, is not entirely unknown in the realm of autonomous driving applications. This entails the modeling of the vehicle's kinematics and dynamics, as well as the surrounding road conditions. Given the unpredictability of actual driving conditions, this can be exceedingly complex, leading many RL methodologies to opt for not considering the specific vehicle model, thus adopting a model-free approach.

Approximate Dynamic Programming (ADP) is a concept fundamentally recognized in Reinforcement Learning (RL), where exploration is integrated into Dynamic Programming (DP) to identify an approximate optimal strategy for addressing problems with large state or action spaces, wherein an exact solution is computationally unviable. Through the approximation of value functions or policies, ADP methods can furnish scalable solutions to the complex challenges encountered in autonomous driving. These methodologies are particularly invaluable in refining control strategies and motion planning algorithms, especially when precise state transition dynamics are either unknown or too complex for explicit modeling.

The most prevalent categorization within model-free Deep Reinforcement Learning (DRL) distinguishes between value-based and policy-based learning approaches. Deep Q-Learning (DQN) integrates deep learning with Q-learning, employing deep neural networks to approximate the Q-function, assessing the value of disparate actions within the current state. This approach enables DQN to manage environments characterized by high-dimensional state spaces, including video games and autonomous driving simulators. The implementation of DQN involves the collection and utilization of data from environmental interactions to train neural networks, where experience replay and target networks serve as crucial stabilizing elements. DQN's capacity to handle complex decision-making processes renders it highly suitable for applications within dynamic and visually rich environments.

Enhancements to the foundational DQN architecture, including Double DQN, Dueling DQN, and the adoption of target networks, have ameliorated challenges associated with training convergence. Three prevalent refinements exist: Double DQN reduces overestimation bias by decoupling action selection from Q-value evaluation. Dueling DQN augments action evaluation by distinguishing between state value and action advantage estimations, furnishing a more detailed comprehension of the impacts of actions. Target networks enhance training stability by moderating the pace of Q-value updates, thereby mitigating fluctuations in learning dynamics.

Policy learning approaches, by facilitating a direct mapping from states to actions, allow agents to ascertain a probability distribution over actions for a given state, diverging from value-based learning approaches which indirectly choose actions to maximize expected returns. Policy Gradient (PG) represents a foundational method within policy learning, aiming to enhance expected returns through the optimization of policy network parameters. This approach is particularly apt for scenarios characterized by continuous action spaces or high dimensions, as seen in robot control and complex decision-making tasks. Policy Gradient facilitates learning by directly modifying policy parameters to amplify the likelihood of beneficial actions and diminish the probability of detrimental actions.

Advanced policy learning algorithms, such as Proximal Policy Optimization (PPO)\cite{schulman2017proximal} and Trust Region Policy Optimization (TRPO)\cite{schulman2015trust}, are designed to tackle the stability and efficiency challenges faced by policy gradient methods in practical implementations. PPO enhances the training process's stability by constraining the magnitude of policy updates, whereas TRPO guarantees updates do not diverge excessively by incorporating a trust region within the optimization process. These techniques offer means to effectively update policies while preserving learning stability, particularly apt for application scenarios necessitating meticulous control over policy update steps.

Value-based approaches face challenges in effectively generating actions within large action spaces, and the training of policy networks is not readily amenable to temporal difference and experience replay methods, resulting in convergence difficulties. The Actor-Critic methodology merges policy and value learning, comprising two models: the Actor, responsible for generating actions, and the Critic, tasked with evaluating the value of these actions. This architecture is designed to harness the advantages of value learning to guide policy learning, enhancing the directionality and efficiency of the learning process. The Actor-Critic approach exhibits distinct advantages when addressing issues involving partially observable or high-dimensional state spaces.

\begin{figure}[t!]
    \centering
    \includegraphics[width=0.6\linewidth]{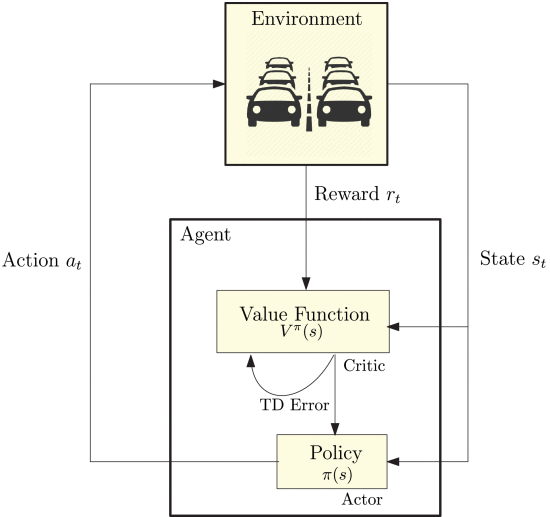}
    \caption{\textbf{Actor Critic method\cite{haydari2020deep}.}}
    \label{fig:AC}
\end{figure}

Enhancements to the Actor-Critic methodology encompass Deep Deterministic Policy Gradient (DDPG)\cite{lillicrap2015continuous}, Twin Delayed DDPG (TD3)\cite{dankwa2019twin}, and Soft Actor-Critic (SAC). DDPG is apt for continuous action spaces, melding the merits of deep learning and policy gradient approaches. The employment of a "deterministic" policy markedly enhances sampling efficiency throughout training, more effectively directing the network's gradient update trajectory. TD3 mitigates overestimation bias via the integration of a twin network structure and delayed update mechanisms, whereas SAC fosters broader exploration of state spaces among agents through entropy regularization, thereby enhancing the stability and robustness of learning.

In the domain of continuous control, reinforcement learning encounters the significant challenge of high-quality sample scarcity, markedly elevating the complexity of training. An efficacious strategy is the adoption of an experience replay mechanism\cite{schaul2015prioritized}, enabling the algorithm to store and subsequently reuse encountered transitions (encompassing states, actions, rewards, and new states) multiple times throughout training. This approach significantly boosts sample utilization efficiency, diminishes learning process variance, and permits batch updating of non-continuous experiences. Building upon this, Deep q-learning from demonstrations (DQfD) \cite{hester2018deep} further broadens the application scope of experience replay, not merely reutilizing experiences generated by the agent itself but also specifically incorporating mechanisms for learning from expert demonstrations. DQfD significantly hastens the learning pace by amalgamating temporal difference updates with the supervised classification of expert actions, even when based on limited demonstration data. Furthermore, through the prioritized replay mechanism, DQfD can adaptively modulate the usage ratio between demonstration data and self-generated data throughout the learning process.

An innovative approach for accelerating training entails initially constructing a policy based on existing data through the supervised learning of the policy network, followed by further optimization and enhancement of this policy via reinforcement learning\cite{zhang2018pretraining}. This approach leverages supervised learning for a rapid initiation of the learning process and employs reinforcement learning for policy refinement, thereby enhancing the efficiency of the learning trajectory. It is particularly well-suited for contexts with ample labeled data, facilitating swift attainment of superior performance levels.

Beyond model-free RL, Model-Based RL considerably enhances training efficiency\cite{kaiser2019model}. Model-Based RL merges the viewpoints of control theory and reinforcement learning, utilizing introduced or fitted environmental models to forecast state transitions, thus facilitating more precise network parameter updates within simulation environments. From the perspective of control, neural networks are employed to fit dynamic models (unknown model calibration), acquiring state transition functions and supplying nominal control strategies for intricate control tasks. From the reinforcement learning viewpoint, policy updates are conducted directly through the fitted environmental model, optimizing objectives with simulator-generated data to boost solution efficiency and diminish the necessity for real-environment interaction. The principal challenge encountered by model-driven RL is that model inaccuracies may induce data biases, particularly over extended periods, where cumulative errors can substantially diminish performance. To tackle this issue, current improvements like Model Value Expansion (MVE) have been developed, limiting the step length of model predictions to curtail cumulative errors\cite{feinberg2018model}. Moreover, strategies incorporating Model-Free RL methods are under investigation to exploit their benefit of direct learning from environmental feedback, alleviating the impact of model inaccuracies on the learning process and enhancing the stability and efficiency of model-driven RL\cite{gu2016continuous}.

\begin{figure}[h!]
    \centering
    \includegraphics[width=0.6\linewidth]{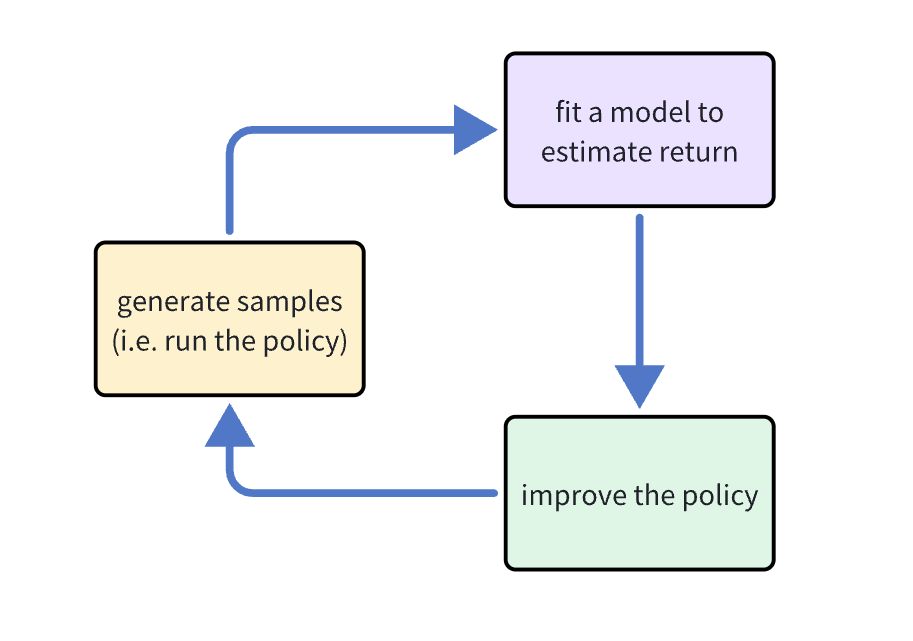}
    \caption{\textbf{ Model-based RL}}
    \label{fig: Model-based RL}
\end{figure}

\subsection{Review of RL-based Autonomous Driving Research}

\begin{figure}[h!]
    \centering
    \includegraphics[width=0.7\linewidth]{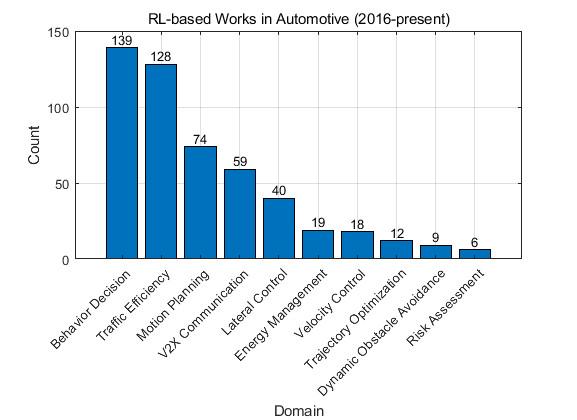}
    \caption{\textbf{ RL-based works in Autonomous Car.}}
    \label{fig: RL-based works}
\end{figure}

In the domain of autonomous driving, the deployment of Reinforcement Learning (RL) is rapidly proliferating, encompassing realms from high-level decision-making to specific motion control. An analysis of literature post-2016 reveals that RL has achieved notable research advancements across various subdomains of autonomous driving. Notably, in the realms of behavioral decision-making (139 papers), traffic efficiency (128 papers), motion planning (comprising 74 papers on motion planning, 12 on trajectory optimization, and 9 on dynamic obstacle avoidance), vehicle-to-everything (V2X) communications (59 papers), and control (encompassing 20 papers on lateral control, 18 on longitudinal control, and 19 on energy management), the volume of papers underscores the research focus and significance of these areas. This illustrates the pivotal role of reinforcement learning techniques in deciphering and enhancing the complex interactions, decision-making processes, and control strategies within autonomous driving systems.

Reinforcement learning is capable of learning optimal strategies through exploration and exploitation mechanisms within unknown or dynamically changing environments, with certain off-policy methods permitting online updates. Given the potential for instantaneous changes in road conditions, traffic flow, and the behavior of surrounding vehicles, autonomous driving systems must be capable of continuously learning from and adapting to new situations. For instance, within the domains of behavioral decision-making and traffic efficiency, RL can aid autonomous driving systems in assessing various potential actions within complex scenarios, selecting the optimal course of action to augment safety and fluidity. In the context of multi-agent systems applications, reinforcement learning has unveiled new perspectives for research within areas like vehicle-to-everything (V2X) communications and traffic flow control. Through multi-agent reinforcement learning, the communication and coordination among vehicles can be optimized, thereby enhancing the efficiency and safety of the entire traffic system. This capability is particularly crucial in highly interactive scenarios, such as intersections and congested roads, where multiple vehicles must make joint decisions to avert conflicts and optimize traffic flow.

Following the analysis of 45 selected recent papers, it has been observed that RL algorithms are extensively applied across diverse subdomains of autonomous driving. Within the context of path planning challenges, algorithms such as SAC, PPO, TD3, DQN, and DDPG are frequently employed, showcasing their robust adaptability in this subdomain. Regarding control challenges, particularly in lateral, longitudinal, and integrated lateral-longitudinal control tasks, the application of the DDPG algorithm is notably pervasive. Simultaneously, model-based DRL and ADP also exhibit their efficacy in these contexts. In the arena of end-to-end control, the application of algorithms tends to be more scattered, attributable to the fact that end-to-end control does not rely exclusively on RL. This reflects a demand for multiple algorithms within complex systems integrating perception with action output.

\begin{figure}[h!]
    \centering
    \includegraphics[width=0.7\linewidth]{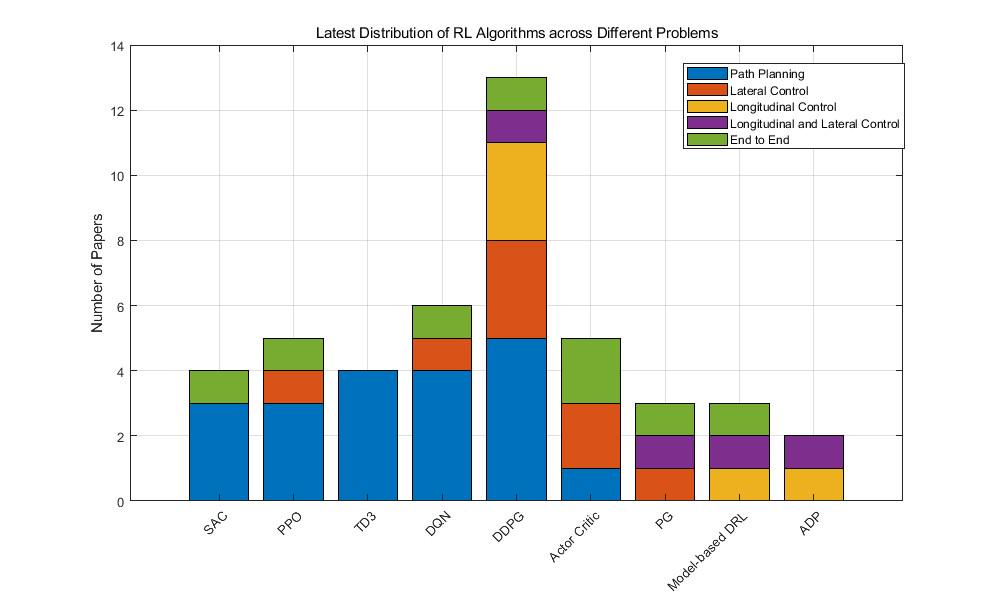}
    \caption{\textbf{ Application of specific RL algorithms to trajectory planning, lateral control, longitudinal control and end-to-end control subproblems.}}
    \label{fig: specific RL algorithms apps}
\end{figure}

Autonomous driving systems encounter a diverse array of complex challenges across various subdomains, with the nature of these challenges dictating the selection diversity of RL algorithms. Within path planning, the pronounced attributes of dynamism and uncertainty render SAC and PPO algorithms, capable of balancing exploration with stability, particularly crucial. These algorithms can effectively navigate the optimal path across fluctuating road conditions while maintaining stability in the learning process. DDPG, by employing its deterministic policy, delivers explicit gradient signals for continuous control issues and diminishes variance throughout the training phase, thus showcasing outstanding performance in continuous control challenges, including smooth acceleration and precise steering. The challenge within end-to-end control revolves around effectively converting high-dimensional perceptual data into immediate action decisions. This necessitates algorithms capable of not only efficiently learning policies but also managing substantial inputs from diverse sensors and devising intricate action sequences.

\section{DRL for Trajectory Planning}

Within autonomous driving systems, trajectory planning is crucial, guaranteeing that vehicles can transition safely and efficiently from their present location to their destination. This chapter will delve into the trajectory planning issue, with a particular emphasis on the application of RL methods in addressing this challenge.

\subsection{Challenges in Trajectory Planning}

The paramount challenge in trajectory planning within autonomous driving systems lies in generating smooth paths that adhere to initial and final conditions as well as an array of constraints. Additionally, algorithms must navigate static and dynamic obstacle avoidance tasks, foresee future vehicle trajectories, and sidestep collisions while tracking arbitrary targets within dynamic environments. Spanning from global to local path planning within intricate environments, prevailing methodologies encompass random search and sampling, curve interpolation, and numerical optimization\cite{gonzalez2015review}. Random search and sampling approaches, notably the Rapidly-exploring Random Tree (RRT) algorithm, furnish an efficacious means for identifying feasible paths within complex terrains. By engaging in random sampling of the configuration or state spaces and examining their connectivity, these strategies facilitate swift planning within high-dimensional spaces.

Paths yielded by RRT might be discontinuous and somewhat rudimentary; however, the RRT* algorithm substantially enhances path optimality via temporal framework optimization, swiftly offering resolutions for intricate driving conditions. To address the discontinuity in paths engendered by sampling approaches, curve interpolation techniques are employed to forge smooth trajectories. Accounting for variables like vehicle dynamics, feasibility, and comfort, these techniques adapt to roadway conditions, generating new datapoints via predefined nodes to guarantee trajectory continuity and satisfy motion constraints. Helical, polynomial, and Bézier curves represent commonly utilized methods capable of consistently delivering high-quality solutions. However, within the context of global path planning, despite the ability to optimize paths and account for road and vehicle limitations, real-time obstacle processing may prove to be considerably time-intensive.

Under dynamically changing environments and constraints, numerical optimization methods manifest their distinct advantages. Through the design of an objective function and its subsequent minimization or maximization, this approach holistically considers vehicle dynamics, environmental data, and constraints in the pursuit of optimal control strategies and paths. The meticulous design of the objective function is crucial for the success of numerical optimization, and within the context of reinforcement learning, it is translated into a reward function, steering the algorithm towards learning the optimal path planning strategy.

Dynamic Programming (DP) and Quadratic Programming (QP) represent two prevalent solution approaches within the domain of numerical optimization. DP utilizes the Bellman optimality principle, iteratively working backwards from the end point to ascertain the optimal path, theoretically capable of identifying the global optimum. However, its computational complexity escalates substantially with the problem size. Quadratic Programming, employed within Model Predictive Control (MPC), involves setting the objective function as the sum of squares of state equations across a series of forecasting periods, underscoring the real-time nature and efficiency of the solution. This renders it more suitable for problems in continuous spaces and dynamic environments. This approach is principally solved via the interior point or active set methods, effectively managing constraints and ensuring that control strategies can be updated in real-time to accommodate dynamic shifts.

The flexibility and adaptability of numerical optimization methods in generating trajectories enable path planning to directly respond to real-time vehicle states and environmental changes. This allows for dynamic adjustments in path and control decisions, taking into account constraints pertaining to roads, vehicles, and other road users. Consequently, the selection of suitable optimization strategies and algorithms, along with the thoughtful design of objective functions, is crucial for enhancing the performance of autonomous driving systems. The two principal challenges confronting current mainstream solution approaches are: first, the vehicle model may be unknown or imprecise, impacting the accuracy of solutions; and second, computational efficiency issues. Particularly, the DP method experiences a drastic increase in computational load when addressing continuous or high-resolution spaces, rendering the solution process exceedingly time-consuming. While MPC mitigates complexity by configuring the objective function as a quadratic form and employing QP, in scenarios necessitating real-time path planning with control loop frequencies demanding up to 100Hz or beyond, QP still necessitates highly optimized algorithms and hardware acceleration to satisfy real-time criteria.

\subsection{Implementing DRL for Trajectory Planning}

In the realm of autonomous driving trajectory planning, while Model Predictive Control (MPC) and Dynamic Programming (DP) methodologies have demonstrated their efficiency and accuracy across numerous scenarios, their adaptability and flexibility are constrained in the face of environmental uncertainties, model inaccuracies, and computational resource limitations. In contrast, Deep Reinforcement Learning (DRL) significantly enhances planning flexibility through its exceptional adaptability to the environment and dynamic learning traits. It generates trajectories via value or policy networks, eschewing reliance on predetermined waypoints in favor of dynamic decision-making based on environmental states and objectives. Concerning trajectory planning challenges, DRL emerges as a promising alternative. We've collated research on DRL's specific applications in this arena into Table 1, spotlighting the scenarios addressed, RL methodologies employed, and experimental approaches undertaken.

In the application of RL to autonomous driving trajectory generation, the design of appropriate states (State) and actions (Action) is crucial, enabling the algorithm to effectively learn and produce optimized trajectories. The survey by Leurent et al.\cite{leurent2018survey} synthesizes various representation methods of states and actions within autonomous driving research. The design of states should encompass the vehicle's current motion status (e.g., position, speed, acceleration, and heading angle), environmental information (e.g., road conditions, traffic signs, and the status of surrounding vehicles), and goal information (e.g., the destination location or planned trajectory), to thoroughly depict the vehicle's interaction with the environment. Additionally, considerations should include lane information, path curvature, the vehicle's past and future trajectories, longitudinal information such as Time to Collision (TTC), and scene information like traffic regulations and the positions of traffic lights. Raw sensor data (e.g., camera images, LiDAR, radar) offers more detailed contextual information, whereas simplified abstract data (e.g., 2D bird's eye views ) reduces the complexity of the state space.

Action design may encompass vehicle control maneuvers, such as steering and speed adjustments, aimed at trajectory generation, or it could involve simplifying the trajectory into an articulation of a sequence of parameters, with actions being those parameters that delineate the trajectory. Prudent design of states and actions must not only encapsulate sufficient information to inform decisions but also sustain efficiency and operability, ensuring that RL algorithms are capable of producing safe and efficient trajectories in real-time and with precision within intricate autonomous driving contexts.

Crafting effective reward functions is vital for the successful deployment of Reinforcement Learning (RL) within autonomous driving trajectory planning. The reward function must comprehensively account for multiple aspects, including the trajectory's safety, smoothness, and speed consistency, encompassing factors such as the distance traveled towards the destination, vehicle speed, maintaining stillness, avoiding collisions with other road users or scene objects, refraining from infractions on sidewalks, remaining within lanes, sustaining comfort and stability amidst avoiding extreme acceleration, braking, or turning, and compliance with traffic laws. The design of reward functions also encompasses advanced techniques like reward shaping\cite{ng1999policy}, encouraging the optimization process to evolve towards the optimal strategy by furnishing the agent with additional, well-crafted rewards. Rewards can also be deduced based on expert demonstrations through Inverse Reinforcement Learning (IRL)\cite{abbeel2004apprenticeship}. In the absence of explicit reward shaping and expert demonstrations, agents may utilize intrinsic rewards or motivations to evaluate the quality of their actions.

In configuring reward functions, it's crucial to ensure that rewards aren't overly sparse, allowing the agent to derive useful feedback from each action. Appropriately setting the weights of reward elements to balance trajectory planning objectives, including safety, efficiency, and comfort, is equally crucial. For scenarios involving urban road networks and intersections, the reward function should account for vehicle behavior within complex traffic settings, encouraging the agent to adhere to traffic regulations and optimize paths through intersections. In emergency avoidance scenarios, there's a particular emphasis required on maintaining safe distances and avoiding collisions, motivating the agent to implement preventive measures. Furthermore, the design of reward functions must contemplate the dynamic shifts in the environment and interactions among multiple vehicles, ensuring the agent can make adaptive choices in fluctuating settings, thereby significantly boosting the robustness and flexibility of trajectory planning.

While Reinforcement Learning (RL) proposes innovative solutions for trajectory planning in autonomous driving, its application encounters numerous challenges. Initially, unlike traditional Model Predictive Control (MPC), RL does not depend on precise environmental models, which proves particularly advantageous when models are inaccurate or entirely absent. However, RL must navigate the constraints of computational resources and the challenges posed by environmental uncertainties and dynamic changes, necessitating algorithms with the capacity for real-time adaptation and updates. Additionally, one of the primary challenges in applying RL to autonomous driving trajectory planning lies in designing a reward structure that can effectively balance multiple objectives and ensuring the solutions' generalizability and adaptability. Specifically, the limited ability to manage emergency scenarios, where most approaches favor pursuing solutions that meet specific rules at the expense of adequately responding to emergencies; the requirement for data preprocessing and dependency on substantial volumes of data exacerbate the challenge of facilitating real-time responses, particularly in emergency situations. These challenges necessitate overcoming through innovative RL methodologies and strategies to realize efficient and safe trajectory planning within complex traffic scenarios.

\subsection{Recent DRL Applications in Trajectory Planning}

\begin{longtable}{@{}p{0.15\linewidth}p{0.1\linewidth}p{0.15\linewidth}p{0.2\linewidth}p{0.3\linewidth}@{}}
\caption{A Comparison of DRL-based Trajectory Planning} \label{tab:table1} \\
\toprule
\textbf{Application Scenario} & \textbf{Work} & \textbf{Algorithm} & \textbf{Experiments} & \textbf{Pros} \\
\midrule
\endfirsthead
\multicolumn{5}{c}%
{{\bfseries Table \thetable\ continued from previous page}} \\
\toprule
\textbf{Application Scenario} & \textbf{Work} & \textbf{Algorithm} & \textbf{Experiments} & \textbf{Pros} \\
\midrule
\endhead
\bottomrule
\endfoot
\bottomrule
\endlastfoot
Intersections & \cite{88} & SAC & Custom-built Simulator & Integration of Reinforcement Learning and Computer Vision \\
Intersections & \cite{92} & PPO & Flow framework & Fully Autonomous Traffic Improvement at Intersections \\
Intersections & \cite{227} & TD3 & SUMO and CARLA & Safety and Efficiency in Multi-Task Intersection Navigation \\
Intersections & \cite{51} & DQN & Custom-built Simulator & Reduction in Fuel Consumption \\
Urban road & \cite{124} & DDPG & CARLA & Success in Unprotected Left Turns \\
Urban road & \cite{570} & DQN & CARLA & Short Training Time \\
Urban road & \cite{581} & TD3 & CARLA and ROS & Improved Convergence and Stability \\
Urban road & {297} & TD3 & CARLA & Hierarchical Method for Object Avoidance \\
Urban road & \cite{53} & Double DQN & Custom-built Simulator & Energy Consumption Reduction in EVs \\
Narrow lane & \cite{122} & TD3 & CARLA & Instantaneous Solution from Pre-trained Network \\
Narrow lane & \cite{195} & Actor critic & Custom-built Simulator And real world in a campus & Negotiation-Aware Motion Planning \\
Highway & \cite{389} & SAC & MATLAB & Increased Security and Efficiency on Highways \\
Racing & \cite{247} & Fuzzy DRL & DeepRacer & Explainable Fuzzy Deep Reinforcement Learning \\
Racing & \cite{207} & TRPO\&PPO & F1Tenth & Residual Policy Learning for High-Speed Racing \\
Racing & \cite{294} & SAC & CARLA & Combining Control Methods for Handling Limits \\
No specific scenario & \cite{223} & DDPG & Custom-built Simulator And Real World in a Campus & Real-time NN-based Motion Planning \\
No specific scenario & \cite{250} & DDPG & real world cases on the ZalaZone & Optimal Vehicle Trajectory Learning \\
No specific scenario & \cite{558} & DDPG & numerical verification & Stability Evaluation with Lyapunov Function \\
Unknown Dynamic Environment & \cite{511} & Dueling Double DQN & ROS and Gazebo & APF-D3QNPER for Superior Generalization \\
Off-road environments & \cite{590} & DDPG & ARL Unity & Covert Navigation in Off-road Environments \\

\end{longtable}

DRL has already showcased its ability to offer solutions within applications targeted at specific scenarios. Path planning challenges are intricately linked to specific application contexts, such as intersections, areas dense with pedestrians, and urban environments with complex traffic flows. The unique attributes of these scenarios are vital for the design of reward functions and experimental validation. Concentrating on these scenarios allows researchers to more precisely simulate real-world challenges and, in turn, formulate effective DRL strategies.

Within the context of intersections, deep reinforcement learning has demonstrated substantial potential in addressing path planning challenges. The complexity inherent in intersection scenarios primarily stems from dynamic traffic flows, multi-vehicle interactions, and the presence of pedestrians, all of which significantly compound the challenges of path planning. In response, multiple studies have employed diverse DRL strategies for mitigation. For instance, Yudin et al.\cite{88} trained intelligent agents for simulated autonomous vehicles through a novel method integrating reinforcement learning with computer vision. Utilizing comprehensive visual information on intersections obtained from aerial photography, they automatically detected the relative positions of all road users and explored the feasibility of estimating vehicle orientation angles via convolutional neural networks. Through the application of modern and efficacious reinforcement learning methodologies like Soft Actor Critic and Rainbow, the convergence of the learning process was expedited by harnessing the acquired additional features. Liu Yuqi et al.\cite{227} proposed a multi-task safety reinforcement learning framework equipped with a social attention module, aimed at enhancing safety and efficiency during interactions with other traffic participants. The social attention module focuses on the statuses of negotiating vehicles, with a safety layer incorporated into the multi-task reinforcement learning framework to ensure safe negotiations.

In urban road network contexts, the complexity is heightened by sudden changes in traffic conditions and the presence of dynamic obstacles, presenting substantial challenges to vehicular safety. Zhou Weitao et al. [124] concentrate on the notably challenging task of unprotected left turns within urban road networks. Utilizing the DDPG algorithm and the CARLA simulator, they introduce an innovative method for trajectory planning. This method not only accounts for dynamically generated trajectory actions but also enhances adaptability and safety through a weighted amalgamation with fixed-strategy trajectory actions. This approach exhibited commendable performance in random and challenging test scenarios, achieving a 94.4 percent success rate in autonomously navigating through 250 intersections. Wang Zhitao et al. \cite{195} utilized an actor-critic methodology to develop a negotiation-aware motion planning framework within a proprietary simulator. This framework, while ensuring vehicular safety and fluidity, takes into account the needs of other traffic participants, adeptly adapting to the complexities of urban traffic, and has undergone real-vehicle validation on campus roads.

Moreover, the most significant challenge posed to reinforcement learning by complex environments is the difficulty associated with achieving training convergence. Clemmons J. and their team\cite{570} significantly reduced training time and enhanced learning efficiency through the application of various optimized DQN algorithms, marking a substantial improvement over most extant image-based approaches. Gao L. et al.\cite{581} utilized the Twin Delayed DDPG (TD3) algorithm to improve the training efficacy of the DDPG network, substantiating the model's convergence speed and stability within CARLA and ROS settings. Cheng Yanqiu et al.\cite{51]} concentrated on longitudinal trajectory planning within mixed traffic flows, employing the Adam optimizer to hasten training. By segmenting vehicle trajectories into portions featuring constant acceleration/deceleration while minimizing assumptions, they circumvented the enumeration of states and actions in value networks within intricate solution spaces, thus boosting the efficiency of model training and inference. Meanwhile, Zhang Ruiqi et al.\cite{207} leveraged TRPO and PPO algorithms within the F1Tenth environment to demonstrate the efficiency benefits conferred upon network training by the residual policy learning method in the context of high-speed autonomous racing scenarios.

\begin{figure}[h!]
    \centering
    \includegraphics[width=0.7\linewidth]{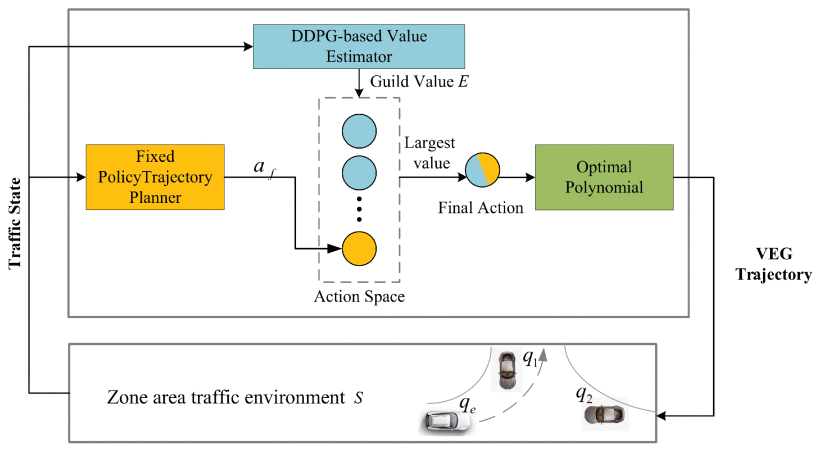}
    \caption{\textbf{ The framework of the Value Estimation Guild (VEG) optimal trajectory planner, a methodology integrating RL with outcomes from rule-based controllers\cite{124}}}
    \label{fig: VEG optimal trajectory planner}
\end{figure}

In scenarios characterized by lane divisions, high-speed driving, and lane-changing maneuvers, the immediacy of decision-making is increasingly crucial. Feher Arpad et al.\cite{122} utilized the Twin Delayed DDPG algorithm to develop a pre-trained neural network and trajectory generation algorithm within the CARLA simulator, capable of delivering solutions within 1 millisecond. This significantly surpasses traditional dynamics models and rule-based methodologies like DP or MPC, making it particularly apt for dynamic trajectory generation tasks. By breaking down the curve of the overtaking process into a sequence of line segments, they substantially simplified the complexity of the issue, enhancing both the real-time nature and applicability of the solution. However, ensuring the stability of the output results amidst varying road conditions and on wet or slippery surfaces continues to pose a challenge.

\begin{figure}[h!]
    \centering
    \includegraphics[width=0.7\linewidth]{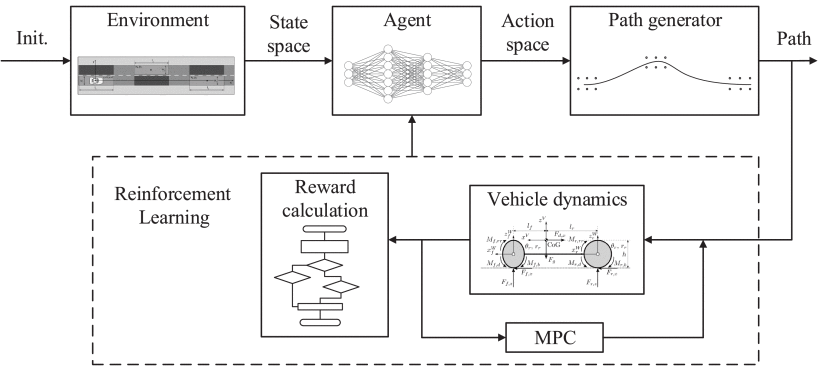}
    \caption{\textbf{ The training and control loop employed in the agent's design. The curve is segmented into eight parts, enabling rapid trajectory generation from a single action output by the agent\cite{122}.}}
    \label{fig: agent trajectory_curve sample}
\end{figure}

Currently, established methods for trajectory generation are predicated on dynamics models and rules. Thus, integrating reinforcement learning into a segment of these traditional methodologies to augment performance constitutes an effective strategy. Zhang Mei et al.\cite{389} employed the SAC algorithm to realize a safer and more efficient path planning solution within the MATLAB environment. By optimizing crucial parameters within the polynomial curve interpolation process, their approach maintained the highest success rates in three and four-lane scenarios, demonstrating how to enhance safety by suitably adjusting driving strategies amidst increasingly complex traffic conditions. Bautista-Montesano et al.\cite{247} enhanced fuzzy logic control utilizing reinforcement learning techniques on the DeepRacer platform, instituting an interpretable Fuzzy Deep Reinforcement Learning methodology, offering an innovative solution for autonomous vehicles. Hou et al.\cite{294} proposed a novel controller amalgamating traditional model-based control, model-free reinforcement learning, and expert knowledge. This controller demonstrated superior strategies and adaptability under extreme vehicle maneuverability boundary conditions, substantiating its excellence and extensibility across diverse complex track conditions.

\begin{figure}[h!]
    \centering
    \includegraphics[width=0.7\linewidth]{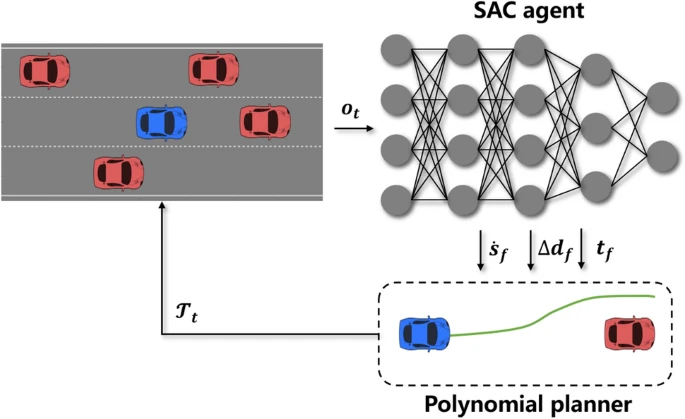}
    \caption{\textbf{ A quintessential approach of incorporating a modicum of ML into traditional solutions, where the agent is employed to generate some key parameters of the trajectory via the curve interpolation method\cite{389}.}}
    \label{fig: agent trajectory ploy-planner}
\end{figure}

Moreover, within the realm of autonomous driving path planning research, the environmental exploration attribute of deep reinforcement learning facilitates its application beyond specific driving scenarios. Existing studies span a broad spectrum, from real-world driving settings to unknown dynamic environments and even off-road conditions. In two studies, Feher Arpad et al.\cite{223}\cite{250} developed a real-time motion planner based on neural networks through the DDPG algorithm, and transitioned it to real-world testing in campus environments and the ZalaZone automobile testing field, underscoring DRL's potential in crafting safe and dependable trajectories. Cabezas-Olivenza Mireya et al.\cite{558} performed numerical validation with the DDPG algorithm, evaluating the navigation stability of agents trained by DDPG through the Lyapunov function, exploring DRL's effectiveness in sustaining navigational stability. Hu Hui et al.\cite{511} in ROS and Gazebo environments, validated the exceptional generalization capabilities of a singular network produced by Dueling Double DQN across diverse scenarios, alongside its superior performance regarding convergence speed, loss values, and path planning duration. In off-road settings, Hossain J. et al.\cite{590} employed DRL to ensure covert navigation while accomplishing low-cost trajectories, demonstrating DRL's potential to tackle complex terrains and unknown obstacles.

\section{DRL for Vehicle Control}

The output of vehicle control is directly connected to specific actuators, constituting the final connection to the physical world within the autonomous driving framework. This chapter will delve into motion control challenges and elucidate the application of RL methods in addressing these issues.

\subsection{Challenges in Path Tracking and Speed Control }

In autonomous driving systems, accurate motion control is essential for achieving safe and dependable navigation. Vehicle motion control can generally be categorized into two principal components: lateral control and longitudinal control. The central objective of lateral control is path tracking, which seeks to ensure that the vehicle precisely adheres to the pre-established path, regardless of whether these paths are straight lines or complex curves. The conventional approach to accomplishing this goal entails initially identifying and modeling the dynamic behavior of the vehicle, selecting essential dynamic variables, and formulating corresponding state equations based on physical laws and real-vehicle testing data. Building upon these foundations, controllers are designed to dynamically adjust the vehicle's driving state in real time, ensuring that the key dynamic variables remain within a specified range\cite{mazzilli2021integrated}, thus following the target path.

For instance, the yaw angle discrepancy between the current vehicle orientation and the desired orientation could serve as the dynamic variable for tracking, given that we can regulate the yaw angle by manipulating wheel steering. Utilizing methods such as look-ahead, corrected look-ahead, or future behavior prediction based on dynamic models, we can ascertain the difference between the current and desired yaw angles. Subsequently, trajectory tracking can be accomplished by controlling steering, traction braking, and other aspects of control quality.

\begin{figure}[h!]
    \centering
    \includegraphics[width=0.7\linewidth]{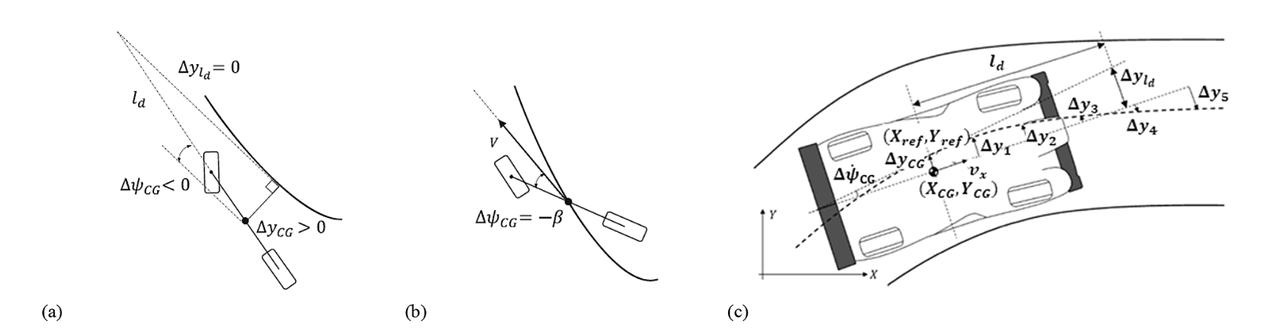}
    \caption{\textbf{ Achieving trajectory tracking by following the yaw angle $\Psi$ \cite{mazzilli2021integrated} based on look-ahead error, (b) based on modified look-ahead error, and (c) through the derivation of a dynamical model.}}
    \label{fig: Follow the angle of yaw ψ}
\end{figure}

In the realm of autonomous vehicle control, speed tracking poses a significant challenge, aiming to maintain or modify vehicle speed to align with a predetermined speed curve. This task is critical for minimizing travel time and ensuring the dynamic stability of the vehicle, aiding in the prevention of rollovers or skidding resulting from excessive speed. The complexity in precisely controlling speed arises from the vehicle's dynamic behaviors, encompassing factors such as the nonlinear interaction between the tires and the road surface and dynamic shifts prompted by alterations in speed and steering angle.

According to a review\cite{stano2023model}, between 2015 and 2021, Model Predictive Control (MPC) emerged as the predominant method in the domain of path tracking control research, constituting approximately 50 percent of the published papers. MPC technology is capable of addressing problems with multiple variables, taking into account constraints on states and control actions, and forecasting future behavior of the system. Other approaches, such as the Linear Quadratic Regulator (LQR), Model-Free Control (MFC), and Nonlinear Model Predictive Control (NLMPC), tackle path tracking and speed control challenges through the adoption of varied strategies. Nevertheless, these methodologies share a common challenge: reconciling the discrepancy between model simplification and the complexities of the real world while ensuring control accuracy. The design and implementation of control strategies become especially intricate in scenarios of high-speed driving and complex road conditions, necessitating the use of more sophisticated models and algorithms to accommodate these variations.

Constructing an accurate vehicle dynamics model is an exceedingly complex endeavor that necessitates the integrated consideration of various factors. These include the intricate interplay between the road surface and tires, alterations in dynamic characteristics with changes in vehicle speed, and the interdependence between lateral and longitudinal dynamics. To simplify this issue, while preserving a certain level of accuracy, simplified models, including kinematic models and dynamic bicycle models, are commonly employed. These models undergo linearization, facilitating easier analysis and implementation during the design of control strategies. However, when vehicles operate at high speeds or encounter complex road conditions, these simplified models may not adequately capture the vehicle's authentic behavior. In such instances, more sophisticated models, like Pacejka's magic formula tire model, are required to enhance the precision of control strategies.

Current control strategies predicated on dynamic models confront several significant challenges: firstly, they necessitate precise system models, but as different vehicles possess disparate parameters, this demands substantial efforts in parameter calibration. Secondly, in certain specific road segments, such as large curves necessitating higher lateral acceleration, the system may falter. Furthermore, the decision-making within these systems typically relies on rule-based definitions, leading to complications in development, the introduction of subjective human factors, and challenges in ensuring comprehensive coverage.

To tackle these intricate challenges, model-free, data-driven methodologies present a novel solution. Contrary to traditional control strategies grounded in precisely defined vehicle dynamics models, data-driven approaches empower vehicles to learn and refine their control strategies based on actual driving data. The application of model-based DRL likewise reveals vast potential. This approach efficaciously narrows the divide between theoretical models and real-world vehicle dynamics. It encompasses not just the theoretical models of vehicle dynamics but also assimilates the dynamic behavior exhibited by vehicles in actual driving scenarios, behavior that is frequently subject to the influence of numerous unpredictable environmental factors.

\subsection{Implementing DRL for Enhanced Vehicle Control}

Implementing augmented vehicle control utilizing DRL enables the RL algorithm to progressively master strategies for adjusting vehicle steering, acceleration, and braking through an ongoing training regimen. This method's advantage lies in its independence from labeled datasets, which facilitates the agent's ability to exhibit robust generalization capabilities across novel scenarios. The training goal is to attain near-optimal trajectory and speed tracking across diverse driving paths, concurrently sustaining the vehicle's dynamic stability. Throughout the learning process, the vehicle dynamically adjusts its state to follow the designated trajectory, optimizing transit time while ensuring not to surpass the boundaries of dynamic stability.

In modeling vehicle motion control issues within a MDP, the state space mirrors that utilized in trajectory planning, encompassing critical factors like vehicle position, speed, acceleration, wheel steering angle, environmental data, and the destination location or planned trajectory. The distinction lies in the control's heightened demand for state dynamism, hence this information is predominantly employed in a simplified and abstracted form.

The action space of an agent consists of control command outputs, typically encompassing modifications to the wheel steering angle, adjustments in acceleration or deceleration, braking, and gear shifting among discrete actuators. The majority of these control commands are continuous values, resulting in a substantially large dimension of the action space. To reduce complexity, the use of DRL algorithms intended solely for discrete action spaces (like DQN) is permissible. Here, the action space can be discretized by segmenting the continuous actuator range into equal-sized intervals\cite{desjardins2011cooperative}. The selection of the number of actuator intervals involves a trade-off aimed at balancing smooth control against the cost of action selection. Employing DRL algorithms to directly learn strategies for managing continuous value actuators (like DDPG) or simplifying the action selection process through a temporal abstraction options framework is viable.

The design of the reward function must encompass multiple objectives, including vehicle stability, efficient driving, and safe obstacle avoidance. Positive rewards are allocated for reaching the target location or maintaining stability, whereas negative rewards are imposed for behaviors like collisions or deviation from the intended path. RL encounters challenges related to computational resource constraints, environmental uncertainty, and dynamic shifts, necessitating real-time adjustments and updates. Although reinforcement learning offers innovative solutions for autonomous vehicle control, its application process is fraught with numerous challenges.

RL does not depend on precise environmental models, an advantage when models are imprecise or absent. Yet, when capable of accurate modeling and necessitating interpretable and readily modifiable control implementations, RL falls short compared to traditional methodologies. A method to enhance the modifiability of RL involves primarily utilizing the existing framework while integrating RL's adaptive and dynamic updating capabilities\cite{247,264}, enabling operation independent of RL's output actions in situations of problems or when identifying emergency constraints. In terms of interpretability, one could endeavor to use a network or model inversely to ascertain the relationship between RL's output actions and input states, thereby attempting to rationalize unforeseen actions undertaken by the agent\cite{217}.

\subsection{Recent DRL Applications in Vehicle Control}

\begin{longtable}{@{}p{0.15\linewidth}p{0.1\linewidth}p{0.15\linewidth}p{0.2\linewidth}p{0.3\linewidth}@{}}
\caption{A Comparison of DRL-based Vehicle Control} \label{tab:summary} \\
\toprule
\textbf{Problem} & \textbf{Work} & \textbf{Algorithm} & \textbf{Experiments} & \textbf{Pros} \\
\midrule
\endfirsthead
\multicolumn{5}{c}%
{{\bfseries Table \thetable\ continued from previous page}} \\
\toprule
\textbf{Problem} & \textbf{Work} & \textbf{Algorithm} & \textbf{Experiments} & \textbf{Pros} \\
\midrule
\endhead
\bottomrule
\endfoot
\bottomrule
\endlastfoot

Path tracking & \cite{57} & DQN and DDPG & CARLA & Realization Process Clarity and DDPG Performance \\
Path tracking & \cite{33} & DQN & Paramics & Travel Time Consistency with Increased Bus Volume \\
Path tracking & \cite{264} & Actor Critic & Driving simulator hardware & Adaptive PID Weight Adjustment \\
Path tracking & \cite{403} & Actor Critic & Driving simulator hardware & SRL-TR2 for Trajectory Tracking \\
Path tracking & \cite{362} & PPO & Numerical Simulation & PPO2-Stanley for Vehicle Tracking and Safety \\
Path tracking & \cite{161} & DDPG & CARLA & DCN-DDPG for Path-Tracking Efficiency \\
Lane following & \cite{175} & DDPG & TORCS & DCPER-DDPG for Lane Following \\
Lateral Control & \cite{337} & Model-based DRL & Numerical Simulation & Model-Based DRL for Lateral Control \\
Lateral Control & \cite{85} & PG & Custom-built Simulator & Monte Carlo Tree Search to reduce complexity of Policy Network \\
Speed Control & \cite{140} & DDPG & MATLAB & DDPG for Multi-Target: Enhanced Safety, Efficiency, and Comfort, VS MPC \\
Speed Control & \cite{196} & Model-based DDPG & Unity & Model-Based DDPG for High-Speed Autonomy \\
Longitudinal Control & \cite{217} & DDPG & Numerical Simulation & Explainable RL for Decision-Making Clarity \\
Integrated Control & \cite{216} & ADP & Numerical Simulation & Two-Phase Data-Driven Policy Iteration \\
Integrated Control & \cite{83} & CNN self attention and RL methods & SUMO & Joint Optimization with CNN Self Attention \\
Integrated Control & \cite{636} & DDPG & Numerical Simulation & DDPG-Enhanced LADRC for 3-DOF Vehicles \\

\end{longtable}

In tackling path tracking and lane-keeping challenges, recent research has made notable advancements across several experimental platforms (e.g., CARLA, Paramics, and TORCS) by employing a range of RL algorithms, including DQN, DDPG, Actor Critic, and PPO. These research findings not only underscore the efficiency and adaptability of RL algorithms in managing complex dynamic systems but also offer fresh perspectives and solutions for enhancing the performance optimization and safety of autonomous vehicles.

Enhancing the generalization capability of models across varied scenarios, especially for effective control within intricate urban traffic contexts, necessitates the comprehensive application of diversified simulation environments, safe exploration mechanisms, model-agnostic control strategies, and real-time adaptive adjustment methods. Through extensive experimentation on simulation platforms like CARLA and Paramics, Perez-Gil et al. \cite{57} and Gao et al. \cite{33} showcased strategies for mitigating risks in practical applications and augmenting the adaptability of algorithms. Hu, Fu, and Wen \cite{337} integrated Dyna-style algorithms with actions derived from Robust Control Barrier Functions (CBF) and employed Gaussian Process (GP) models to boost sample efficiency, facilitating effective learning in uncertain environments while ensuring safety. Wang, Zheng, Sun \cite{403}, by combining ADRC with DDPG, not only exploited ADRC's model-agnostic property for estimating and compensating for uncertainties and external disturbances but also facilitated real-time adaptive adjustment of control parameters via DDPG-optimized strategies, thereby further improving generalization capabilities and execution efficiency in variable environments.

\begin{figure}[h!]
    \centering
    \includegraphics[width=0.7\linewidth]{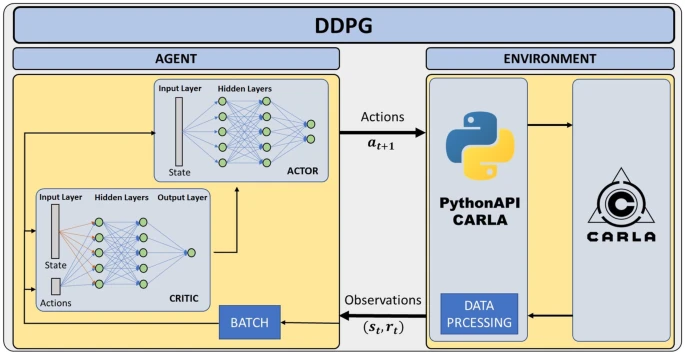}
    \caption{\textbf{  DDPG-based DRL controller architecture\cite{57}}}
    \label{fig: DDPG-based controller}
\end{figure}

For tasks requiring higher precision control, without compromising safety and stability, the efficacy of model-free approaches seldom exceeds that of methods grounded on precise models, suggesting that model-based RL might hold greater advantages. Chen et al. \cite{140} determined speed control rewards on rough surfaces by configuring a combination of rewards for safety, efficiency, and comfort, subsequently employing a DDPG-based trained speed control model to foster safe, efficient, and comfortable vehicle following behaviors. They reported that the trained network exhibited lower overall metrics for speed, acceleration, and clearance distance compared to MPC control. However, their MPC methodology did not disclose the time step, and the objective function merely assigned all weights as 1, suggesting room for refinement. Moreover, the model employed was inferred from a kinematic model without detailing the kinematic process\cite{zhu2020safe}. Despite comparative studies, the upper limit of control precision achievable by model-free RL approaches remains an aspect warranting further contemplation.

On another note, Hartmann et al.\cite{196} devised a model-based DRL approach targeted at the time-optimal speed control problem, formulating time-optimal speed control strategies and employing numerical solutions to anticipate and mitigate scenarios potentially causing vehicle instability. This methodology was validated within the Unity environment, substantiating its adept fit to the dynamics model, outperforming singular model predictions after a training duration of five minutes. Moreover, it achieved superior control outcomes compared to model-free approaches. Similarly, the work of Cui et al.\cite{216}, employing a data-driven approach, enhanced the deductive process of dynamic planning grounded in dynamic models. The implementation of Adaptive Dynamic Programming (ADP) elevated the precision of trajectory tracking across diverse road surfaces.

\begin{figure}[h!]
    \centering
    \includegraphics[width=0.7\linewidth]{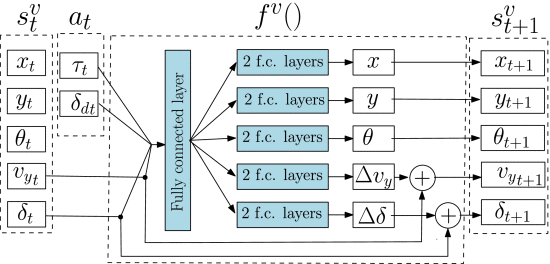}
    \caption{\textbf{  Utilize a moderately simple network to approximate the state transition function\cite{196}.}}
    \label{fig: model-based treans func}
\end{figure}

In high-curvature sections or emergency obstacle avoidance scenarios, significant variations in path curvature present substantial challenges to maintaining path tracking accuracy. He et al.\cite{175}'s methodology, through refinements to the Deep Deterministic Policy Gradient (DDPG) algorithm involving the integration of a dual critic network and a prioritized experience replay mechanism, has enhanced the algorithm's stability and accuracy in navigating high-curvature and complex road conditions. Incorporating extensive high-curvature and complex road scenarios within the simulation environment enables this approach to not only preclude trajectory tracking failures but also to ensure elevated tracking precision. The enhanced DDPG algorithm not only improves lane-tracking performance but also bolsters the model's adaptability to a variety of driving conditions. Luo et al. \cite{362} implemented a hybrid control strategy integrating the robotic Stanley trajectory tracking algorithm with Deep Reinforcement Learning (DRL) technology. Leveraging the efficient path tracking capability of the Stanley algorithm along with the self-learning and adaptability features of DRL, this strategy not only enhances tracking accuracy but also seamlessly integrates collision avoidance capabilities, thus facilitating safe control in emergency situations.

In the context of vehicle control, action spaces are typically continuous, necessitating more complex networks for finer control, a factor that frequently complicates training convergence. Numerous studies have focused on accelerating convergence speed and enhancing convergence stability during training. Yao et al. \cite{161} and He et al.\cite{175} enhanced the DDPG algorithm by incorporating a dual critic network and a prioritized experience replay mechanism, addressing numerous deficiencies inherent to the original algorithm and thereby improving training efficiency and accuracy; Kovari et al. \cite{85}, in a method analogous to that of Alpha Go, leveraged the combination of Monte Carlo tree search with short-term reward strategy networks. This approach necessitates focusing solely on immediate rewards during the training of the strategy network, substantially diminishing the network's complexity and computational requirements, thereby facilitating rapid convergence and real-time operation. Addressing the dichotomy between Q-learning training complexity and the granularity of action spaces, \cite{zhang2018human} applied the Double DQN methodology atop a framework of finely discretized action spaces, incorporating real-world driving data. This approach simulates the real-world commuting experience, facilitating the validation of the devised vehicle speed control system within a simulated environment.

As outlined in the preceding discourse, faced with the complexity inherent to managing multiple chassis control subsystems and potential subsystem inter-conflicts, Deep Reinforcement Learning (DRL) is poised to facilitate integrated motion control. This entails orchestrating steering, traction, and braking actuators’ outputs in unison to address lateral, longitudinal, and postural control issues effectively. Cui et al. \cite{216} introduced a bi-phase data-driven policy iteration algorithm designed for achieving optimal longitudinal and lateral control in autonomous vehicles. This algorithm ensures stable convergence towards a suboptimal controller solution, notably independent of trailing vehicle dynamics, a claim substantiated through numerical simulations. Hu et al. \cite{337} utilized model-based Deep Reinforcement Learning (DRL) to facilitate learning from scratch in real-world conditions, leading to tangible deployment. This illustrates DRL's practical applicability and remarkable adaptability in autonomous vehicle control scenarios. Chen et al. \cite{83} leveraged CNNs, self-attention networks, and deep reinforcement learning approaches to realize the joint optimization of perception, decision-making, and motion control. Validated within the SUMO simulation environment, this research underscores the potential of integrated optimization control strategies. Wang et al. \cite{636} demonstrated a DDPG-augmented Linear Active Disturbance Rejection Control (LADRC) controller applied to three degrees of freedom (3-DOF) autonomous vehicles, evidencing significant adaptability to uncertainties and enhanced control performance through real-time parameter tuning facilitated by deep reinforcement learning. While Reinforcement Learning (RL) has seen some application in integrating control across various sub-problems, it is crucial to acknowledge that for essential safety-critical control subsystems, such as ABS and ESP, RL's near-term replacement remains challenging due to insufficient temporal and empirical validation.

Considering the transition of models trained within simulated environments to real-world contexts, the direct linkage of vehicle motion control to the physical realm necessitates a cautious approach when employing data-driven methods with limited interpretability. Wang et al. \cite{403} devised a practical framework utilizing an actor-critic model. Under the auspices of safety constraints, they developed an RL-based trajectory tracking system and subsequently implemented this system as a lateral controller for full-sized vehicles. The crux of their approach was the employment of a lightweight adapter to forge a mapping between simulated environments and the real world. This strategy underwent validation on driving simulation hardware, affirmatively testing the model’s efficacy in actual driving situations, thereby facilitating a smoother transition from simulated frameworks to tangible applications.

\section{Integrated Motion Planning and Control via DRL}

The process of human driving is inherently an end-to-end system. While drivers’ brains may process intermediate outcomes like predicting the actions of other road users, selecting driving maneuvers, and local trajectory planning, these processes lack the explicit interfaces characteristic of layered decision-making controls. Humans are capable of navigating vehicles almost solely with visual cues. End-to-end supervised learning approaches strive to emulate human visual-dependent driving by directly mapping input sensor data to vehicular control commands.

\subsection{End-to-End DRL Methods for Autonomous Driving }

End-to-end applications within the vehicle control sector have showcased their proficiency in effectively realizing well-defined goals. \cite{song2023reaching} employed Reinforcement Learning (RL) for drone control, utilizing a policy network comprised of two layers of multilayer perceptrons for local path planning and motion control. This method concentrates directly on enabling the drone to traverse specific points, bypassing the generation of detailed path trajectories. This approach facilitated ultra-high-speed control, outperforming the premier human contestants in aerial races. Additionally, this study explored drone control via optimal control techniques; however, the pivotal reason for RL's superiority over optimal control lies in its ability to "offer improved optimization objectives," essentially direct effect realization.

\begin{figure}[h!]
    \centering
    \includegraphics[width=0.7\linewidth]{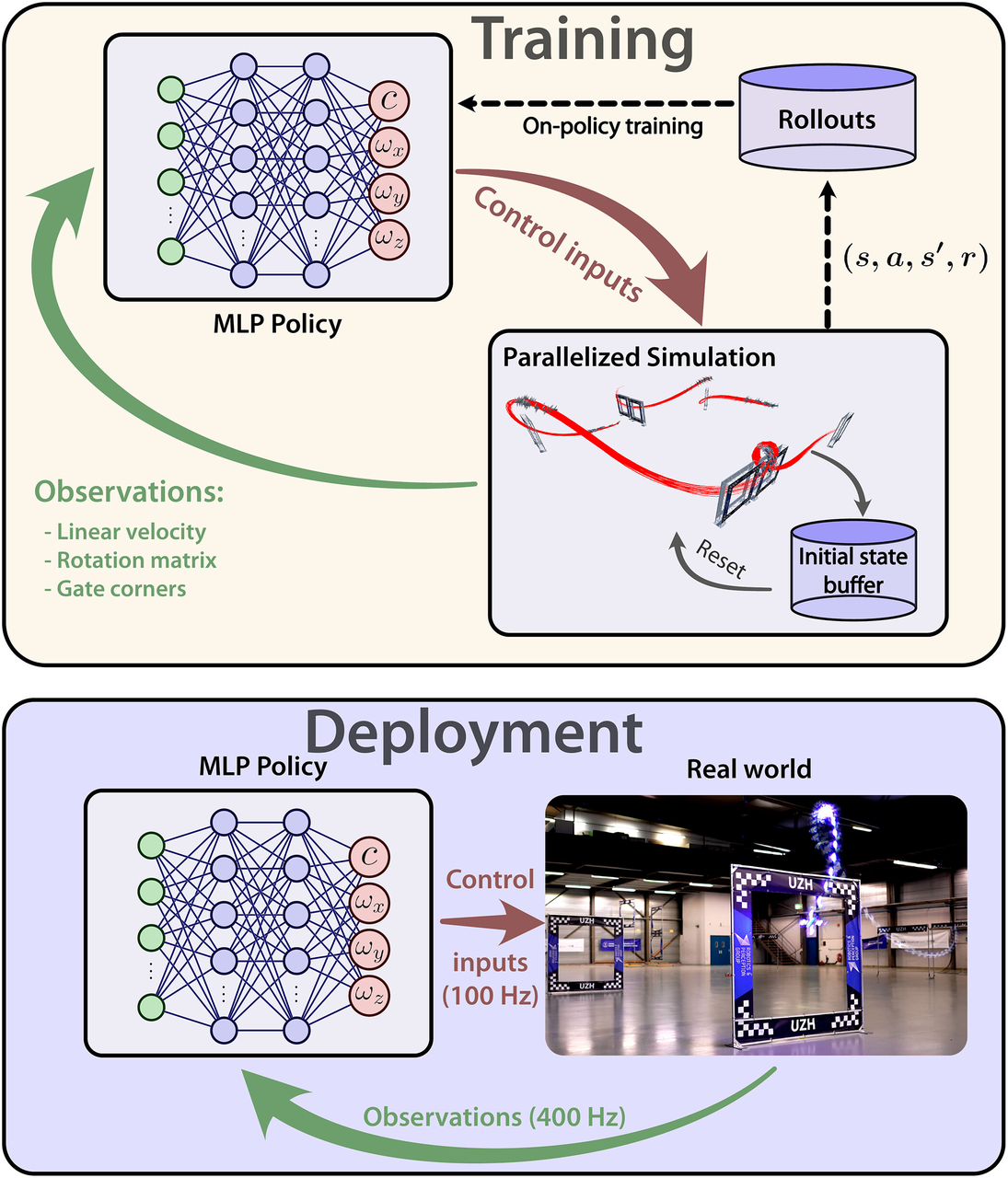}
    \caption{\textbf{ End-to-end control approaches in drone racing\cite{song2023reaching}.}}
    \label{fig: end2end_UV}
\end{figure}

Beyond goals directly tied to control mechanisms, extant studies have elucidated how the intricate interplay between decision-making, planning, and execution facilitates enhanced control performance. Chen et al.\cite{83} engaged in the joint learning and optimization of three critical components of autonomous vehicles—perception, decision-making, and motion control—a facet frequently overlooked in prevailing autonomous driving research. They utilized a Deep Reinforcement Learning (DRL) methodology, particularly through the development of a novel state representation mechanism. This mechanism processes sensory data via attention and convolutional neural network (CNN) layers, thereby augmenting the overarching efficacy of the autonomous driving strategy. Sensory data is initially funneled through attention layers, focusing on the extraction of pivotal local information, subsequently processed by CNN layers to incorporate a comprehensive view of global information, thus achieving a superior representation. The collaborative learning of decision-making and motion control contemplates the symbiotic relationship between these modules to forge a more efficacious autonomous driving strategy, a claim substantiated within the SUMO simulation environment.

\begin{figure}[h!]
    \centering
    \includegraphics[width=0.7\linewidth]{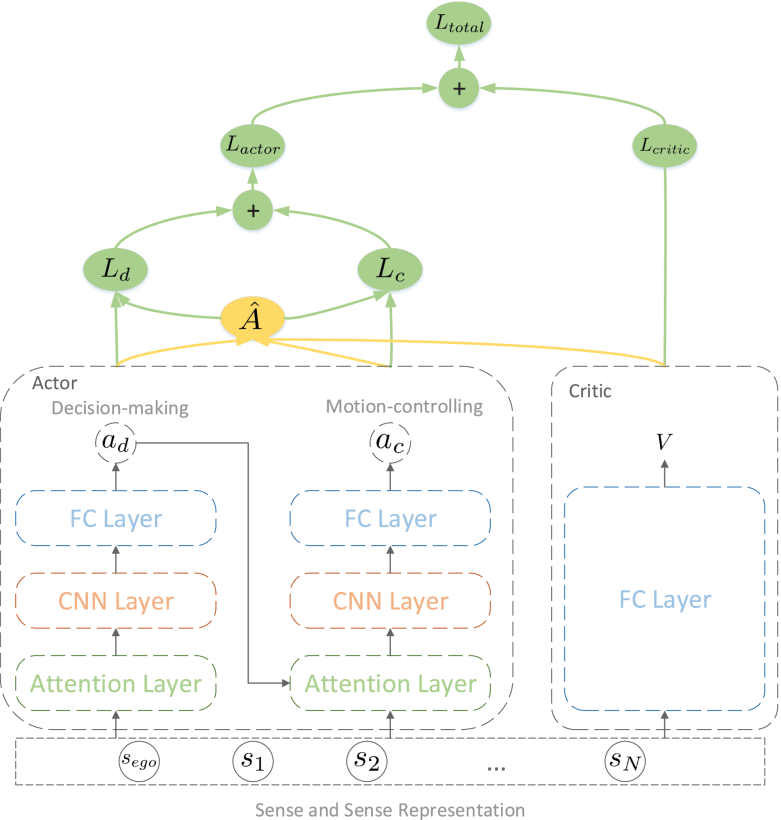}
    \caption{\textbf{The overall system architecture to tactfully processes the sensing data for improvement of control \cite{83}.}}
    \label{fig: processes the sensing data for improvement of control}
\end{figure}

A strategy for realizing end-to-end control involves leveraging supervised learning, capitalizing on extensive sensor and control output data. Long Short-Term Memory networks (LSTMs) are particularly apt for intricate sequence prediction tasks, attributed to their efficacy in managing extended sequence data, discerning long-term dependencies, and circumventing gradient vanishing or exploding issues, a capability reflected across a myriad of end-to-end applications. L. Chen et al. introduced an innovative motion planning framework dubbed "Parallel Planning," integrating artificial traffic scenarios with deep learning models. This framework empowers autonomous vehicles to assimilate environmental cues and adeptly navigate emergencies in a manner akin to human drivers, consequently elevating autonomous driving's safety and sophistication. S, N.F. et al. unveiled an LSTM-based end-to-end model tailored for applications involving autonomous vehicles merging onto highways. Employing driving simulator hardware for both training and evaluation phases, this research harnessed expert driver data to instruct the LSTM network in executing high-speed merging maneuvers both accurately and securely.

The advent of large-scale models has prompted researchers to favor the development of increasingly complex models trained on extensive datasets. In online applications, end-to-end networks are capable of directly deriving necessary driving maneuvers from sensor outputs, thereby enhancing the coherence and immediacy of decision-making control. Nonetheless, this methodology is not without its significant shortcomings. Primarily, the vast dimensionality of state spaces in autonomous driving tasks necessitates data in exponential volumes, culminating in a profound need for extensive datasets. Presently available open-source datasets, including Waymo Open \cite{llc2019waymo}, Oxford Robotcar \cite{maddern20171}, ApolloScape \cite{huang2019apolloscape}, Udacity \cite{rivard2013udacity}, and ETH Pedestrian \cite{du2019self}, predominantly furnish data related to environmental perception, object detection, and tracking but do not directly supply driving behavior annotations. Additionally, given this strategy's intrinsic reliance on neural networks' capacity to generalize across data points, its safety assurances become challenging to validate in the face of novel or infrequent data instances.

\subsection{Implementations of End-to-End DRL in Driving Automation}

\begin{longtable}{p{0.1\linewidth}p{0.2\linewidth}p{0.25\linewidth}p{0.3\linewidth}}
\caption{Deep Reinforcement Learning Applications in Autonomous Driving} \label{tab:DRLApplications} \\
\toprule
\textbf{Work} & \textbf{RL Algorithm} & \textbf{Experiments} & \textbf{Pros} \\
\midrule
\endfirsthead
\multicolumn{4}{c}%
{{\bfseries Table \thetable\ continued from previous page}} \\
\toprule
\textbf{Work} & \textbf{RL Algorithm} & \textbf{Experiments} & \textbf{Pros} \\
\midrule
\endhead
\bottomrule
\endfoot
\bottomrule
\endlastfoot

\cite{wang2022end}& PPO & CARLA & PPO with GAE, wild environment \\
\cite{6} & PG & Real-World & VISTA for Real-world Policy Transfer \\
\cite{355} & SAC & Real-World & SESR for Enhanced Interpretability in Sim2Real \\
\cite{627} & LSTM & Driving simulator hardware & LSTM for Highway Merging \\
\cite{486} & A3C & Custom-built Simulator And real word in a campus & A3C for Automated Lane Changing \\
\cite{guan2022integrated} & Model-based RL & Custom-built Simulator And real word & IDC Framework, real-world, Optimized Control Excellence \\

\end{longtable}

Contrasting with end-to-end controls predicated on supervised learning, Reinforcement Learning (RL) boasts the capability to gather data within simulated environments, facilitating expedited training across various contexts. The studies conducted by Amini et al.\cite{6} and Chung et al.\cite{355} have illuminated the efficacy of end-to-end RL techniques within the realm of autonomous driving applications. Utilizing the VISTA data-driven simulator, Amini et al. adeptly transitioned policies honed through RL to authentic driving contexts, navigating challenges presented by unfamiliar roads and intricate near-collision scenarios, thereby showcasing the bolstered robustness of end-to-end RL strategies in intricate navigation tasks. This substantiates the premise that strategies cultivated within simulators can efficaciously generalize to real-world thoroughfares, adeptly managing unprecedented scenes and complex predicaments, thus unveiling the potential of employing RL for effective perception and robust operations. Concurrently, the SESR method introduced by Chung et al., leveraging category-decoupled latent encoding, not only augmented the interpretability of end-to-end autonomous driving systems but also significantly alleviated the simulation-to-reality (Sim2Real) distribution shift issue, further validating the successful deployment of end-to-end RL approaches in genuine environments and their contribution to the refinement of control maneuvers.

\begin{figure}[h!]
    \centering
    \includegraphics[width=0.7\linewidth]{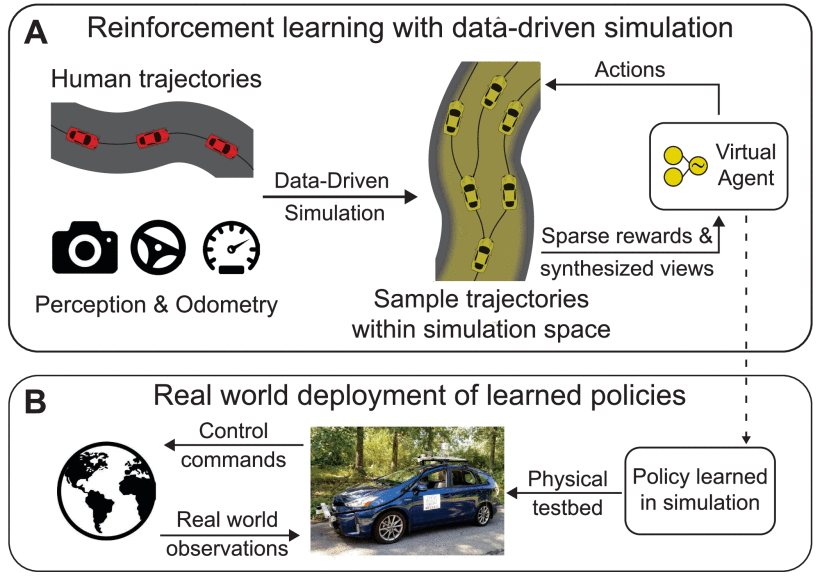}
    \caption{\textbf{The process of deploying an end-to-end method. (A) Initiates with pre-training via supervised learning, subsequently undergoing RL optimization within the VISTA simulator (B) Transitions to deployment in the real world\cite{6}.}}
    \label{fig: end2end_VISTA}
\end{figure}

Analogous to trajectory planning endeavors, contemporary end-to-end Reinforcement Learning (RL) techniques are predominantly confined to particular scenarios. This limitation chiefly arises from the dependency on state and reward function designs intricately linked with specific scenarios, a condition that, even within singular tasks, presents substantial complexity and necessitates extensive manual fine-tuning. Addressing lane-changing dilemmas, Zhou et al.\cite{486} experimented with an end-to-end Asynchronous Advantage Actor-Critic (A3C)-based RL framework, boosting exploration efficacy via a multi-threaded setting, and secured stable convergence through weighted averaging methodologies. Wang et al.\cite{wang2022end} executed a lane-keeping feature on the CARLA simulation platform, utilizing images for input and translating these into steering maneuvers and acceleration outputs. A pioneering end-to-end deep reinforcement learning model, predicated on the Proximal Policy Optimization (PPO) algorithm, was devised expressly for the autonomous navigation of off-road Unmanned Ground Vehicles (UGVs). This model amalgamates various functionalities facilitated by PPO, encompassing gradient computation, objective functions, and amendment mechanisms, tailored to particular application scenarios and demonstrated to surpass the Soft Actor-Critic (SAC) approach in simulated settings. Furthermore, this model utilizes the Generalized Advantage Estimation (GAE) algorithm to diminish training variability and streamline the hyperparameter optimization process, rendering it optimally suited for scenarios characterized by high-dimensional states.

In another investigation, the Integrated Decision and Control (IDC) framework introduced by Y Guan exhibited a model-based reinforcement learning approach that synergizes seamlessly with optimal control techniques. This methodology adeptly navigates the real-world application challenges of securing precise models and managing high computational complexities, which traditional strategies often encounter. Building upon a meticulously trained vehicle fitting model, this technique bifurcates the driving task into static path planning and dynamic optimal tracking. In contexts where computational speed demands are moderate, it employs optimal control strategies to forge statically viable paths, whereas actual execution relies on policy networks for path selection and adherence. This approach was validated within simulated environments featuring four-lane highways and intersections, followed by the agent's transition to real-world implementations. It showcased remarkable online computational efficiency and extensive applicability.

\begin{figure}[h!]
    \centering
    \includegraphics[width=0.7\linewidth]{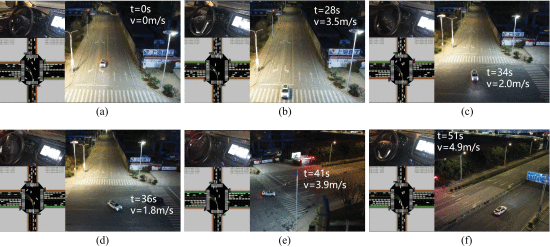}
    \caption{\textbf{ The real-world testing of the devised IDC framework. (a) Halting and awaiting the green traffic light. (b) Accelerating towards and entering the intersection. (c) Slowing down to prevent a collision and altering the route. (d) Accelerating to advance first upon selecting a safer pathway. 	(e) Establishing a set route and following it. (f) Successfully navigating through the intersection.\cite{guan2022integrated}}}
    \label{fig: end2end_realworld}
\end{figure}

\section{Challenges and Future directions}

\subsection{Enhancing Training Stability and Convergence }

Within the domain of autonomous vehicle planning and control, the stability and convergence of reinforcement learning model training represent a pivotal challenge. Elevating model complexity is intended to bolster adaptability to intricate scenarios and generalization capacity; however, it concurrently escalates the requisite for substantial volumes of high-quality training samples, amplifying the complexity inherent in the learning trajectory. Enhancements to the training regimen primarily concentrate on two dimensions: sample quality and algorithmic refinement.

Within domains like autonomous driving and robotic control, amassing high-quality training samples grapples with hurdles related to latency, reward scarcity, and the uneven dispersion of observational outcomes across vast state spaces. Notably, when accruing valuable insights is either prohibitively expensive or fraught with risk, sample efficiency emerges as a pronounced difficulty. At present, numerous studies concentrate on refining methods for the acquisition of training samples. This includes investigations into the enhancement of training quality through the introduction of noise into training samples\cite{eberhard2022pink}, alongside research aimed at boosting sample efficiency via the definition of the Bellman Eluder dimension\cite{jin2021bellman}. Collectively, these approaches strive to optimize sample usage to bolster training efficiency.

Concurrently, numerous enhancements to reinforcement learning algorithms are designed to augment the stability and hasten the convergence rate of the training regimen. For instance, the Soft Actor-Critic algorithm incorporates entropy regularization into policy optimization to mediate between exploration and exploitation. Meanwhile, the Twin Delayed Deep Deterministic policy gradient (TD3) algorithm mitigates estimation biases and overestimation concerns by integrating dual value networks alongside policy smoothing methodologies. Moreover, model-based RL approaches enhance the stability and acceleration of training through the development of environmental models and the utilization of forecasts regarding dynamic environmental shifts.

Utilizing pre-training and transfer learning strategies can significantly alleviate the challenges and costs associated with obtaining real-world road data, while expediting the convergence of models. Moreover, meticulously crafted reward functions can mitigate issues related to sparse rewards, recalibrate reward magnitudes, and augment the coherence and explicability of rewards, furnishing stable and efficacious direction throughout the training phase. Recent studies have explored the utilization of offline reinforcement learning for the pre-training of generic agents. Notably, approaches that expand offline reinforcement learning to employ extant static datasets for the training of value functions have showcased the capability for swiftly adapting to novel tasks and hastening the online learning of new task variations\cite{macaluso2024small}.

\subsection{Addressing Comparability Issues }

Within the sphere of DRL application research, conducting comparisons across disparate methodologies and their performance outcomes in varied simulation settings constitutes a crucial aspect of the investigative process. Nevertheless, DRL is significantly reliant on the caliber of code execution and meticulous adjustments of hyperparameters. Should there be deficiencies in the code or inappropriate hyperparameter configurations, algorithms that are theoretically efficacious may falter in practical deployment scenarios. This article straightforwardly discusses the adjustment of hyperparameters, code refinement, and the amalgamation of various optimization strategies, culminating in performance that substantially outstrips that of other DQN methodologies in Atari games.

The DRL training regimen is subject to numerous uncertainties, encompassing environmental variability, methods of initialization, mechanisms for action selection, and strategies for experience replay. Variations in random seeds can also significantly impact the quality of results and the efficacy of convergence. Owing to these intrinsic attributes of DRL, directly comparing disparate models poses substantial challenges. Moreover, the comparison of model performances across varied testing platforms is profoundly influenced by the inherent properties of the data, including distribution heterogeneity, scale, and quality. In the absence of unified evaluation criteria and benchmarking tests, such comparisons may culminate in deceptive conclusions.

Systematized approaches to hyperparameter searching, including grid search, random search, and Bayesian optimization, alongside parameterization strategies that sustain activation scale consistency throughout training, exemplified by µTransfer technology, significantly curtail the arbitrariness and subjectivity associated with manual tuning. This, in turn, facilitates the standardization of hyperparameter optimization and network architecture formulation. In their work on end-to-end off-road driving tasks, Wang et al.\cite{wang2022end} utilized the Generalized Advantage Estimation (GAE) algorithm to diminish the volatility of PPO training and streamline the hyperparameter tuning process, achieving a superior overall return relative to outcomes yielded by the SAC methodology. They further hypothesized that this advantage is likely attributable to the enhanced efficacy of GAE in the context of hyperparameter optimization for PPO.

Investigating training and validation frameworks for RL facilitates equitable comparisons among diverse methodologies. The integration of standardized evaluation procedures and performance metrics, coupled with benchmark testing grounded in open-source scenario repositories, guarantees that varied approaches are juxtaposed under identical conditions, thereby cultivating a harmonized assessment framework. The disclosure of comprehensive details regarding experimental configurations, code implementations, and hyperparameter settings, alongside the employment of a standardized data reporting format, aids fellow researchers in replicating and corroborating experimental findings, thereby augmenting the transparency and reliability of comparative analyses.

Beyond juxtaposing various DRL algorithms, numerous articles benchmark their network outcomes against established control methodologies like Dynamic Programming (DP) and Model Predictive Control (MPC). These comparisons endeavor to showcase DRL's capacity to outperform conventional control strategies in designated tasks, furnishing pivotal substantiation for DRL's broader deployment in real-world application scenarios. Nonetheless, it's critical to acknowledge that within the realm of vehicle control, devising and executing methodologies such as MPC presents considerable challenges. Several investigations, despite employing numerical simulations or developing simulated environments, have not thoroughly validated the potential of established control strategies. Comparisons of this nature frequently overlook variances in the implementation conditions of diverse methods, potentially culminating in inequitable assessment outcomes. Consequently, it is advised that such research endeavors should enhance the quality of mature control method implementations within benchmarks, or solicit participation from vehicle control sector practitioners, to guarantee the fairness and precision of these comparisons.

\subsection{Bridging the Simulation-to-Real-World Gap}

Transitioning from simulated to real-world environments represents a dynamic area of inquiry within Deep Reinforcement Learning (DRL), driven by simulations serving as extensive, accurately annotated, and economically viable data reservoirs. Via domain adaptation on both feature and pixel dimensions, OpenAI\cite{andrychowicz2020learning} adeptly trained a robotic arm within the GYM environment, enabling it to execute grasping maneuvers in the real world sans supplemental real-world training. This endeavor hinges on the Domain Randomization strategy, which entails integrating randomness into the simulation training milieu to encompass potential real-world scenarios, thus narrowing the divide between simulated and actual environments. Within autonomous driving, research has involved training A3C agents in environments featuring simulation-to-real\cite{pan2017virtual} converted imagery, subsequently assessing their performance against real-world driving datasets.

Yet, transitioning from simulated environments to the real world constitutes one of the foremost challenges encountered by DRL within the sphere of autonomous driving applications. Although simulated settings offer a safe and controllable learning context, the real world's complexity and unpredictability vastly surpass those of their simulated counterparts. To guarantee a model's performance and stability under authentic road conditions, an array of targeted technological approaches and strategies must be employed. The crux involves enhancing the fidelity of simulation models, necessitating the creation of simulated environments grounded on accurate GIS and real-time traffic data, in conjunction with deploying sophisticated sensor data for the meticulous modeling of environmental dynamics.

The efficacy of transfer strategies is profoundly contingent upon the caliber and methodologies employed in data processing. Optimal training datasets ought to encompass a broad spectrum of driving scenarios, such as severe weather conditions, diverse traffic configurations, and myriad emergency situations, to affirm the model's adaptability to the real world's intricacies. Concurrently, navigating the uncertainty of data and harmonizing datasets from varied origins and frequencies are critical for affirming the efficacy of simulation training.

Conversely, while our present inquiries into generalization predominantly focus on the adaptability to diverse scenarios, attention must also be granted to agents' capacity for interfacing with varying hardware, thereby enhancing an agent's integration with its "body".  The application of sophisticated agents within the actual physical realm offers numerous referential methodologies within the domain of embodied intelligence. This applicability extends not solely to autonomous vehicles but also to realms encompassing industrial unmanned vehicles and robots equipped with mobile platforms, representing a research direction of considerable value.

Beyond the preparation of models and data, actual deployment necessitates considerations of ethics, safety, and compliance with regulations, rendering real-world testing a complex endeavor. Augmenting the validation capabilities of simulation environments presents a viable solution. Recent investigations \cite{feng2023dense} have navigated the bottlenecks of safety verification via Dense Deep Reinforcement Learning (D2RL) techniques. The adoption of intelligent testing environments and the study of adversarial maneuvers have enhanced the resilience and dependability of autonomous driving technologies against the complexities of the real world.

\subsection{Improving Computational Efficiency and Real-time Processing }

During the model evaluation stage, assessing the model's accuracy is crucial, but evaluating its real-time performance—specifically, whether the model's response time and processing speed satisfy the real-time demands of autonomous driving—is equally important, particularly upon incorporating complex architectures like the Transformer. To surmount this challenge, the employment of model optimization techniques—such as model pruning, quantization, and knowledge distillation—serves as a pivotal strategy for enhancing computational efficiency and real-time processing prowess. By diminishing the complexity and computational requisites of the model, these techniques facilitate an expedited inference process. Additionally, the design of lightweight network structures and the utilization of hardware acceleration technologies—like GPUs and TPUs—represent effective measures to bolster the real-time performance of autonomous driving systems.

Model deployment represents a pivotal phase, entailing challenges in transitioning models from development settings (like Python) to real-world applications (such as C++ inference frameworks). Prominent inference frameworks, including TensorRT, OpenVINO, and TVM, are designed to optimize model execution efficiency on specified hardware platforms, facilitating efficient model conversion and deployment. While these frameworks aid in the efficient operation of models on vehicular computing platforms, the deployment process must still navigate challenges related to performance disparities, environmental compatibility, real-time demands, and constraints on resources. A holistic strategy is requisite to guarantee that models not only adhere to the stringent real-time criteria of autonomous driving but also maintain stable operation within environments constrained by resources.

\subsection{Safety}

In \cite{garcia2015comprehensive}, two methodologies for training safety-compliant agents were outlined: incorporating a safety layer into the decision-making process or modifying the optimization criteria. Within the realm of DRL's application to autonomous driving, a safety layer ensures operational safety by real-time monitoring of potential risks and intervening when potential violations of safety constraints are detected. The safety layer, as an integral part of the system, not only perpetually monitors and assesses potential safety risks but also corrects or substitutes strategies as needed to avert dangerous scenarios. As an instance, Gu et al.\cite{397} detailed a state-based method for augmenting safety in autonomous driving, which notably enhances both performance and safety within intricate highway contexts by amalgamating dynamic goal setting with adaptive safety constraints via hierarchical reinforcement learning.

In supervised learning, meticulously adjusting input data serves as an efficacious method to enhance the precision of model outputs. Within reinforcement learning, implementing hard constraints to curtail agents from undertaking perilous actions could precipitate sparse rewards issues, complicating the convergence of the learning process. In reaction, an indirect restriction on agents' outputs through the preprocessing of input data might be viable, as opposed to directly limiting their action selection. State elements that precipitate hazardous actions can be pinpointed through feature engineering and state filtering techniques. The fundamental principle behind this strategy is to optimize input data in a manner that steers agents away from potentially hazardous or superfluous actions, without detracting from the agent's thorough and profound exploration of the environment. It's pertinent to acknowledge that excessively constraining the input could diminish the agent's exploratory capabilities. Consequently, an approach of indirectly restricting outputs could be employed either during the pre-training stage of the agent or upon the training attaining a certain threshold. This entails preprocessing input data to navigate agents away from perilous or undesired behaviors, thus harmonizing the imperatives of safety and exploratory capacity.

\section{Conclusions}

This article provides a thorough review of recent literature, exploring the application of deep reinforcement learning in unmanned vehicle path planning and control. The document encapsulates the application of DRL across specific contexts, including intersection management, urban road navigation, and highway motoring, whilst elaborately discussing how these technologies tackle pivotal autonomous driving challenges like dynamic obstacle detection, adaptability to traffic conditions, and environmental generalization. Despite the presence of established examples within these domains, the incorporation of DRL has notably enhanced the operational efficacy and decision-making optimization of autonomous driving systems.

An analysis was conducted on the application of diverse DRL methodologies within pivotal subdomains of autonomous driving, encompassing path planning, vehicle control, and end-to-end control. Within the realm of path planning tasks, SAC, PPO, TD3, DQN, and DDPG exhibited exceptional prowess, notably in scenarios necessitating dynamic adaptation and intricate decision-making processes. For control tasks, DDPG has been extensively utilized owing to its explicit gradient signaling and elevated sample efficiency in addressing continuous control challenges. Model-based DRL and ADP have manifested their worth in long-haul control endeavors that demand accurate dynamic modeling and control. End-to-end control is marked by its complexity and is frequently employed alongside self-attention mechanisms, optimal control strategies, and model-based RL, to name a few. Such integrative technologies proficiently navigate the hurdles associated with end-to-end learning.

As hardware computational capabilities enhance and algorithms are refined, the deployment of DRL in autonomous driving is set to broaden, accompanied by increasingly complex neural network architectures. The discourse underscored the challenges associated with enhancing the comparability, stability, generalization capacity, and post-deployment vehicular inference efficiency, alongside the imperative of facilitating transitions to real-world applications and ensuring safety during unforeseen circumstances. To aptly showcase DRL's efficacy within autonomous driving, the development of more efficient and stable training methodologies, more precise environmental simulation techniques, and stricter safety norms is requisite. The amalgamation of these technological strides and practical implementations harbors the potential to realize autonomous vehicles that are smarter, safer, and more efficient.

\section*{Acknowledgments}
This work is xxxx. This research was partially supported by the xxxxx.

\bibliographystyle{unsrt}
\bibliography{ref}

\end{document}